\documentclass[10pt]{article}
\usepackage[margin=1in]{geometry}
\usepackage{times}
\usepackage{natbib}
\usepackage{amsmath}
\usepackage{amssymb}
\usepackage{graphicx}
\usepackage{hyperref}
\usepackage{url}
\usepackage{xcolor}
\usepackage{placeins}
\usepackage{booktabs}
\usepackage{array}
\usepackage{microtype}
\hypersetup{hidelinks}

\title{TerrainForge: Physics-Grounded Road Geometry Editing for Counterfactual Autonomous Driving}

\author{Yang Chen\textsuperscript{1} \and Yicheng Zhu\textsuperscript{1} \and Zhenning Li\textsuperscript{2} \and Tao Li\textsuperscript{3} \and Zilin Bian\textsuperscript{1,*}\\[3pt]
\textsuperscript{1}Rochester Institute of Technology \quad
\textsuperscript{2}University of Macau \quad
\textsuperscript{3}City University of Hong Kong\\[2pt]
\textsuperscript{*}Corresponding author}
\date{}

\begin{document}
\maketitle

\begin{abstract}

Road geometry (e.g., crests, sags, and speed humps) and surface conditions (e.g., wet or icy pavement) affect how vehicles move, what drivers and onboard cameras observe, and how much clearance remains between vehicles. Editing these properties in a driving scene therefore requires corresponding changes in vehicle motion. Capturing these differences in a driving video requires a road edit to propagate to vehicle motion, camera viewpoint, and the clearance between vehicles. We present \textbf{TerrainForge}, a framework for generating road geometry-focused counterfactuals from reconstructed multi-vehicle driving episodes. A unified road model connects scene deformation with four-wheel vehicle dynamics, allowing crests, sags, speed humps, and friction changes to propagate through vehicle motion, camera viewpoint, and inter-vehicle clearance. Vehicle dynamics are evaluated against CarSim, and prescribed road geometry is verified in reconstructed Waymo scenes. Across 18 episodes, leaving surrounding vehicles on their recorded trajectories instead of recomputing their responses produces median peak differences in predicted ego-lead distance of 1.52 m for crests and 1.41 m for sags. We further simulate the ego response to 15,758 road edits across 983 braking episodes, pairing each edit with its safety outcomes relative to an unedited replay. These pairs train a first-stage screening surrogate that takes the original driving context and candidate road-edit parameters as input and predicts the resulting change in the ego's terminal gap. On held-out scenes, this prediction achieves $22$--$40\%$ lower mean absolute error than predicting no change, so candidates can be screened cheaply before the full multi-vehicle rollout.

\end{abstract}

\section{Introduction}

\begin{figure}[t]
\centering
\includegraphics[width=\linewidth]{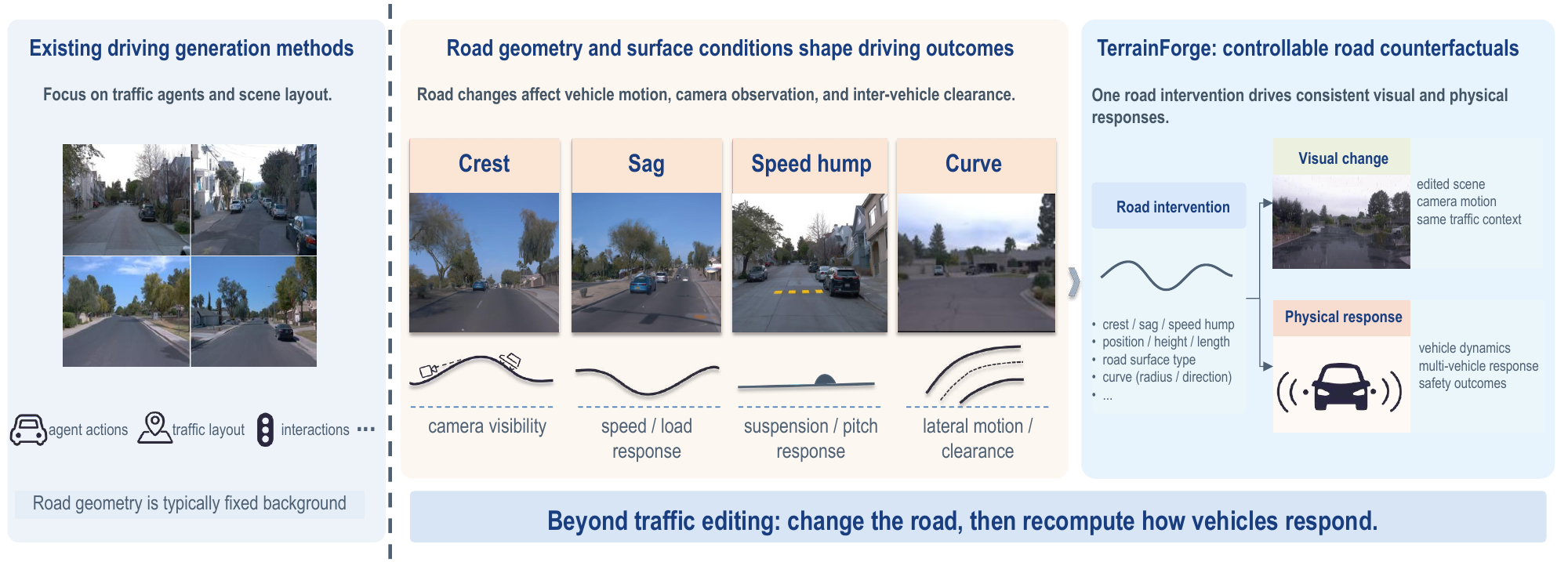}
\caption{TerrainForge edits the road while preserving traffic context, propagating each intervention to scene geometry, vehicle motion, and inter-vehicle clearance.}
\label{fig:concept}
\end{figure}

A vehicle braking behind a lead car can stop with different margins depending on the road beneath it, even when its control inputs remain unchanged. A crest or sag changes road grade and tire loading, while a speed hump excites the suspension and reduced surface friction limits the forces available for braking. These effects alter vehicle motion and camera viewpoint. Vehicles encountering the same road feature at different speeds or times may respond differently, changing the clearance between them. For autonomous-driving evaluation, such cases motivate a controlled comparison: change the road geometry or surface conditions in a recorded episode and examine how the vehicles and their observations change (Figure~\ref{fig:concept}).

Controllable driving generation provides considerable flexibility in scene appearance, road layouts, and traffic configurations. Road maps, vehicle layouts, camera parameters, and text can specify the content of generated scenes\citep{magicdrive2024,panacea2024,geniedrive2026}. Risk-aware trajectory prediction provides a complementary view of how future vehicle motion and driving behavior can be modeled\citep{bian2025strap,bian2024less}. Reconstruction-based approaches have also begun to connect scene editing with explicit models of vehicle physics\citep{roves2026,musyncgs2026}. The question considered here concerns the consequences of a road change across a recorded traffic episode: how the vehicles sharing that road respond, how their relative motion changes, and how those changes appear in the resulting video.

Answering this question requires the edited road to determine the physical conditions experienced by each vehicle. Road shape affects wheel contact and loading, while surface friction constrains the tire forces that produce motion. The resulting changes in position and attitude must then be reflected in the rendered scene. Surrounding traffic is part of this calculation. Leaving neighboring vehicles on their recorded trajectories would combine ego motion produced on the edited road with traffic motion produced on the original road. The estimated clearance would consequently depend on which vehicles were allowed to respond to the edit.

We introduce TerrainForge, a framework for constructing these road counterfactuals from reconstructed driving episodes. A unified road model represents elevation, gradient, and tire–road friction, accommodating continuous crest and sag curves, local features such as speed humps, and changes in surface condition. The geometry specifies how the Gaussian scene is deformed; the geometry and friction together supply road inputs to a four-wheel model with coupled longitudinal, lateral, and vertical dynamics. Each tracked vehicle retains its initial state and the control sequence inferred from its recorded motion while its trajectory is recomputed on the edited road. Updated vehicle poses and camera motion then drive rendering. Holding these inputs fixed allows paired simulations to isolate the road-induced changes under the vehicle model.

Evaluation covers both the accuracy of the edits and their effects on traffic. Vehicle responses are compared with CarSim, recovered road profiles are checked against real-road observations, and prescribed geometry is verified in reconstructed Waymo scenes. The multi-vehicle experiments show that surrounding vehicles’ responses affect the estimated consequences of an edit. Across 18 episodes with usable ego-lead pairs, retaining the recorded motion of surrounding traffic produces median peak ego-lead gap discrepancy of 1.52 m for crests and 1.41 m for sags relative to recomputing its physical response. Accounting for the road’s effect on the ego vehicle alone can therefore give a different assessment of the clearance between vehicles.

The resulting intervention–response pairs also provide supervision for predicting the effects of candidate road changes. These effects depend on the traffic encountering the edited region: the same road feature may have different consequences at different speeds, vehicle separations, or braking strengths. A bank of 15,758 road edits across 983 braking episodes, in which only the ego is re-simulated, supports surrogate models that take the original driving context and road-edit parameters as input. The models predict changes in terminal gap relative to an unedited replay and identify edits with material effects on safety-related outcomes. On held-out scenes, terminal-gap prediction achieves $22$--$40\%$ lower mean absolute error than predicting no change. The surrogate is a first-stage ego-response screen: it prioritizes candidates, and the full multi-vehicle rollout gives the final result.
Our main contributions are:

\begin{itemize}

\item We introduce TerrainForge, a physics-grounded
real-to-counterfactual framework that treats road geometry as a controllable
intervention in recorded driving scenes. By preserving the recorded traffic
context and recovered vehicle controls while changing the road, TerrainForge
generates paired counterfactual episodes for road edits such as crests, sags,
and speed humps.

\item We develop a unified road model that drives both
visual and physical changes from the same intervention. The same road field
governs Gaussian-scene deformation, camera motion, and coupled four-wheel
vehicle dynamics, allowing the edited terrain and the responses of multiple
vehicles to remain physically synchronized.

\item A scene-disjoint intervention–response dataset and first-stage surrogate models that predict the ego's terminal-gap change and identify consequential road edits, screening candidates cheaply before the full multi-vehicle rollout.

\end{itemize}

\begin{figure}[t]
    \centering
    \includegraphics[width=\linewidth,trim=6 9 5 55,clip]{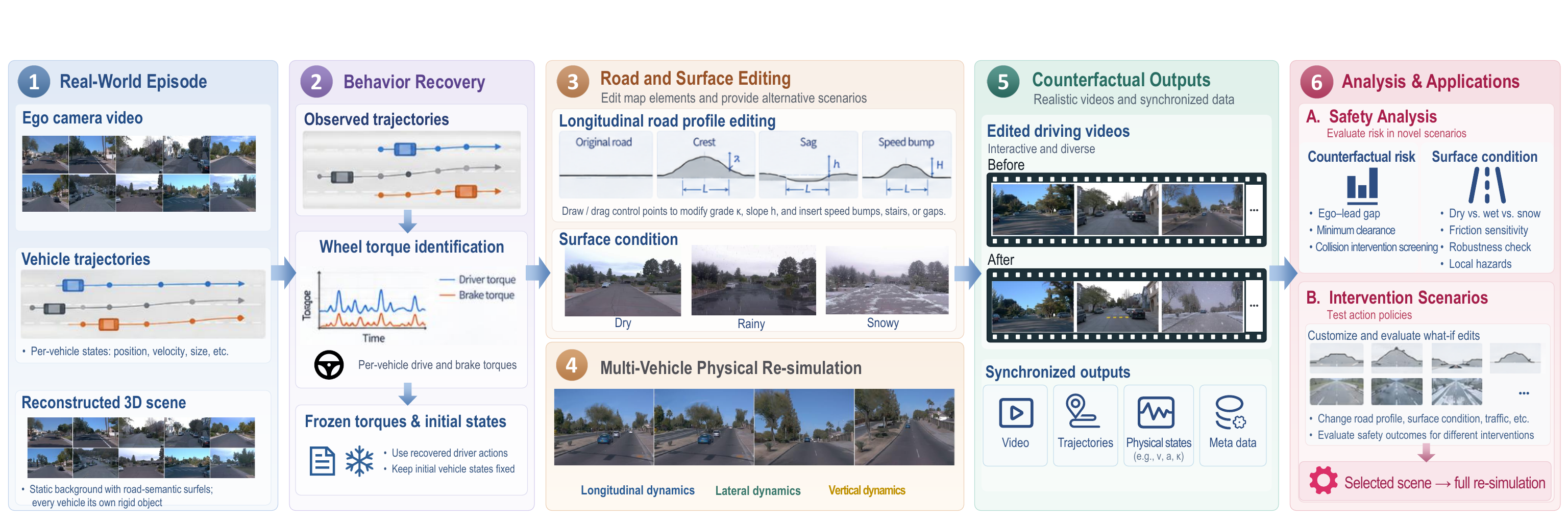}
 \caption{\textbf{Overview of TerrainForge.}
TerrainForge converts a reconstructed real-world driving episode into an
editable, physics-grounded counterfactual environment by recovering
per-vehicle driving actions and allowing interventions to road geometry and
surface conditions. Each intervention jointly determines the reconstructed
scene geometry and the physical excitation experienced by all vehicles.
Replaying the recovered vehicle controls on the intervened road produces synchronized vehicle trajectories, camera observations, rendered videos, and physical states, enabling paired analysis of road-induced changes in motion and inter-vehicle safety outcomes.}
    \label{fig:pipeline}
\end{figure}

\section{Related Work}
\paragraph{Controllable driving-world generation.}
Controllable driving-world generation provides scalable and diverse scenarios for autonomous-driving training, simulation, and evaluation. Existing methods condition generation on HD maps, BEV layouts, 3D boxes, occupancy, and other structured scene representations to control traffic participants and road structure~\citep{gaia1,panacea2024,magicdrive2024,drivingdiffusion2024,uniscene2025,scenariodreamer2025,roadweaver2026}. Recent methods further support substantial changes to road structure, including substantial changes to structured road layouts and rare scene configurations~\citep{drivepts2026}. TerrainForge instead treats road modification as a geometric intervention rather than a visual condition. These controls primarily specify plan-view scene structure and desired vehicle motion. A shared road field jointly drives reconstructed geometry and vehicle response while the remaining conditions of the recorded episode are held fixed.

\paragraph{Physics-aware driving-world generation}
Recent driving-world models improve physical plausibility by incorporating trajectories, motion priors, occupancy, simulator states, or other structured conditions into image and video generation~\citep{drivephysica2024,geniedrive2026,phygenesis2026,mad2026}. These methods mainly emphasize geometric and kinematic consistency, such as plausible object motion, trajectory following, and temporal coherence, while the physical response of a vehicle to changes in road geometry or surface conditions is rarely modeled explicitly.

Reconstruction-based approaches further couple photorealistic 3D scenes with explicit simulators or physics models to support physically consistent motion, interaction, and scene editing~\citep{carlags2026,real2sim2026,roves2026}. muSync-GS couples road elevation and weather-dependent friction to ego-vehicle braking, wheel slip, normal loads, pitch, and camera motion~\citep{musyncgs2026}. TerrainForge formulates road editing over a recorded multi-vehicle episode using a continuous unified road model that captures grades, crest and sag curves, local terrain features, and surface friction. Vehicle-specific controls are recovered and held fixed while the shared road field drives coupled longitudinal, lateral, and vertical dynamics together with scene deformation and re-rendering.

\paragraph{Vehicle dynamics and multi-vehicle simulation}

Physics-based driving simulators and dedicated vehicle-dynamics tools model tire--road interaction, suspension response, road geometry, and coupled longitudinal and lateral motion, while recent platforms increasingly support large-scale multi-vehicle simulation~\citep{carla2017,waymax2023,gpudrive2024,scenefactory2026,carsim2026}. Video accident anticipation and safety-potential-field methods provide related risk-assessment perspectives for autonomous driving~\citep{bian2024accident,bian2025cspf}. These systems provide rich physical responses but generally operate in simulator-native environments rather than preserving the appearance and scene-specific context of a recorded real-world drive. \textsc{TerrainForge} instead embeds continuous road interventions and vehicle dynamics into a reconstructed real-world episode, preserving its visual context while re-rendering the resulting vehicle motion.
\section{Method}
\subsection{Overview and counterfactual formulation}

TerrainForge takes a reconstructed driving episode, calibrated camera poses,
map geometry, and tracked vehicles as input. A road intervention is represented by
\begin{equation}
\mathcal{R}_{\eta}(x,y)=
[z_{\eta}(x,y),\nabla z_{\eta}(x,y),\mu_{\eta}(x,y)],
\label{eq:road_field}
\end{equation}
where $z_{\eta}$ is elevation and $\mu_{\eta}$ is tire--road friction.
For vehicle $a$, the counterfactual rollout is
$\tau_{\eta}^{a}=\Phi(x_{0}^{a},U^{a},\mathcal{R}_{\eta})$,
with initial state $x_0^a$ and control input $U^a$. Within each paired
comparison, initial states, controls, timestamps, and horizon are fixed while
only the declared road variables change.

Our primary setting replays controls inferred from the recorded motion.
Prescribed-control and speed-tracking variants are used only for the
secondary experiments listed in Table~\ref{tab:setup}.

\subsection{Unified road model}
\label{sec:road_model}

TerrainForge defines a single road-intervention field in the metric world frame of each episode. The same spatial parameterization and intervention are used for vehicle dynamics and Gaussian-scene deformation, while the resulting ego-vehicle motion determines the counterfactual camera trajectory.

A reference route in the world frame ($+z$ up) defines road-aligned coordinates $(s,d,h)$, where $s$ is arc length, $d$ is lateral offset, and $h$ is height. A world position is mapped to the road coordinates by $(s,d)=\pi(x,y)$. The baseline profile $H_0(s)$ is obtained from the smoothed ego-route elevation in reconstructed scenes and from the HD-map lane center in the Waymo Motion episodes of Section~\ref{sec:screening}. All reported experiments parameterize a road-geometry intervention as an increment $\Delta H_\eta(s)$ on the native surface:
\begin{equation}
z_{\eta}(x,y)
=
z_{0}(x,y)
+
w(d)\Delta H_{\eta}(s),
\qquad
(s,d)=\pi(x,y),
\label{eq:profile_delta}
\end{equation}
where $w(d)$ controls the lateral support of the intervention and is set to one across the edited road in our experiments. An absolute target profile $H^\star$ can equivalently be represented by $\Delta H_\eta=H^\star-H_0$. The vertical curves and compact features used as $\Delta H_\eta$ are defined in Appendix~\ref{app:road_profiles}.

The shared quantities across geometry and dynamics are therefore the projection $\pi$ and the intervention field. At each integration step, every wheel queries $z_\eta$, $\nabla z_\eta$, and $\mu_\eta$ at its world position, while each affected Gaussian is displaced using the same $\pi$ and $\Delta H_\eta$. Thus corresponding locations in the reconstructed scene and the vehicle model receive the same prescribed road change. Vehicle responses depend on the grade and curvature of $z_\eta$, and both
representations receive the same $\Delta H_\eta$ at the same locations, so each can
keep its own native elevation without altering the intervention
(Appendix~\ref{app:geometry}).

Because $\mathcal{R}_\eta$ is defined in world coordinates rather than in a vehicle-local frame, all vehicles query the same physical road regardless of travel direction. The grade experienced along a vehicle path $\mathbf{p}(q)$ follows from the directional derivative of the shared surface along that path; consequently, vehicles traversing the same intervention from opposite directions experience the corresponding reversed longitudinal grade while remaining coupled to the same world-fixed geometry.
\paragraph{Surface condition.}
Friction is the third component of $\mathcal{R}_{\eta}$ and lives in the same frame (a base coefficient with optional patches; Appendix~\ref{app:method}). It is an authored variable declared for each study: factual grounding (Section~\ref{sec:grounding}) is run on the source surface and the frozen controls are then replayed on $\mu_{\eta}$, so a friction edit is the change in outcome of identical controls.

\subsection{Factual Grounding and Coupled Vehicle Rollout}
\label{sec:grounding}
Waymo provides tracked box poses but no pedal or steering signals, so each vehicle's controls are inferred from its recorded motion on the source road $\mathcal{R}_{0}$, with one fixed procedure across scenes.

\paragraph{Longitudinal control.} Let $q_{k}$ be the recorded path distance of the box center at frame $k$ ($10\,\mathrm{Hz}$). Finite differences of tracked centers reach tens of $g$, so the speed target is a Savitzky--Golay derivative, $\hat v_{k}=\max\!\big(0,\,\mathrm{SG}_{11,2}[q]'_{k}\big)$ (window $11$ frames, i.e.\ $1.1\,\mathrm{s}$, quadratic), and a track is split wherever the implied speed exceeds $40\,\mathrm{m/s}$, which marks a label jump rather than motion. Only this target is smoothed; the replayed path keeps the raw track. A leading stationary interval ($\hat v<0.5\,\mathrm{m/s}$) is held at the recorded pose and the rollout starts at launch. The longitudinal force demand is initialized on the source road,
\begin{equation}
F^{(0)}(t)=m\,\dot{\hat v}(t)+mg\sin\!\big(\arctan H_{0}'(q(t))\big),
\label{eq:ilc_seed}
\end{equation}
distributed to the wheels as $T_{i}=R\,\rho_{i}F$ with a fixed split $\rho=(0.276,0.276,0.224,0.224)$ and $|T_{i}|\le 4000\,\mathrm{N\,m}$, and refined by iterative learning against the model's own rollout on $\mathcal{R}_{0}$, $v^{(j)}=\Phi_{v}(x_{0},T^{(j)},\mathcal{R}_{0})$:
\begin{equation}
F^{(j+1)}=F^{(j)}+\lambda\,m\,\frac{\mathrm{d}}{\mathrm{d}t}\big(\hat v-v^{(j)}\big),\qquad \lambda=0.8,
\label{eq:ilc}
\end{equation}
for at most six iterations or until the speed RMSE falls below $0.03\,\mathrm{m/s}$, keeping the best iterate. The resulting torques $U^{a}=\{T_{i}(t)\}$ are frozen.

\paragraph{Path.} In multi-vehicle episodes each vehicle drives its own unrolled road, whose station is its recorded path distance and whose height is the alignment pulled back along that path, so crossing and oncoming vehicles are not forced onto the ego station. Steering is zero in this frame and the plan-view path is replayed: these experiments exercise longitudinal and vertical coupling, not lateral. The single-vehicle banks (Sections~\ref{sec:lateral} and~\ref{sec:screening}) instead re-solve planar motion, with front steering $\delta_{f}=\arctan\!\big(L\,r/\max(v_{x},0.5\,\mathrm{m/s})\big)$ from the smoothed yaw rate $r$ and a non-iterative inverse-dynamics torque (Appendix~\ref{app:method}).

\paragraph{Counterfactual.} The baseline and edited runs start from the same state and replay the same $U^{a}$, and the reported trajectory is the recording displaced by their difference,
\begin{equation}
\tau^{a}_{\eta}=\tau^{a}_{\rm rec}\oplus\big[\Phi(x^{a}_{0},U^{a},\mathcal{R}_{\eta})-\Phi(x^{a}_{0},U^{a},\mathcal{R}_{0})\big],
\label{eq:difference}
\end{equation}
where $\oplus$ moves the recorded pose along its path by the difference in path distance and adds the differences in heave, pitch, and roll. A zero edit is therefore an exact identity, and identification residuals cancel to first order. Factual replay error measures identification quality on $\mathcal{R}_{0}$; counterfactual accuracy is assessed against CarSim under identical inputs (Section~\ref{sec:carsim_validation}).

\paragraph{Vehicle model.}
The 32-state coupled vehicle model includes planar motion, heave, pitch, roll,
four unsprung masses, tire-slip dynamics, and wheel rotation.
Each wheel queries road height, gradient, and friction; suspension and tire
contact determine normal load, while TMeasy provides longitudinal and lateral
forces subject to the friction circle
$\sqrt{F_{x,i}^2+F_{y,i}^2}\leq\mu_iF_{z,i}$.
Full equations and initialization are given in Appendix~\ref{app:method}.
\subsection{Physics-synchronized scene rendering}
\label{sec:rendering}
\paragraph{Scene and camera.}
Road and sidewalk Gaussians are displaced by the shared road intervention and
rotated to the edited surface normal, while supported objects follow their
footing and unrelated geometry remains fixed. The ego rollout supplies the
counterfactual camera pose through the fixed vehicle-to-camera calibration.
The deformed Gaussian scene is then re-rendered from these poses rather than
warping recorded frames. Detailed attachment and deformation rules are given
in Appendix~\ref{app:scene_deformation}.
\paragraph{Traffic participants}
\label{sec:traffic_objects}

Moving tracked vehicles with sufficient visual support are reconstructed once
as rigid object-centric Gaussian models. Each vehicle is placed according to
its counterfactual metric pose obtained from the edited-minus-baseline rollout,
and all reconstructed vehicles are rasterized jointly with the deformed
background, so occlusion follows scene depth.

Rendering-only contact and depth corrections align vehicles with the
reconstructed road surface. These corrections do not affect evaluation:
clearance, collision, and terminal gap are computed from the metric
counterfactual poses. Reconstruction and placement details are in Appendix~\ref{app:method}.
\subsection{Surrogate prediction}
\label{sec:surrogate}\label{sec:intervention}

The surrogate is a first-stage screen in a two-stage procedure: it ranks candidate edits
from ego-only rollouts, and the selected candidates are then evaluated by the full
multi-vehicle rollout, which gives the final result. Each paired ego rollout provides an
intervention--response sample. Given episode $e$ and edit $\eta$, the surrogate maps pre-simulation
features $\mathbf{x}(e,\eta)$ to terminal-gap change $\Delta G$,
minimum-clearance change $\Delta d_{\min}$, and a consequential-edit
probability.

The input contains four groups of pre-simulation features: ego--target
kinematics, native-road context, intervention parameters, and closed-form
physics features. We use histogram gradient-boosted regressors for the two
continuous targets and a class-balanced boosted-tree classifier for
consequential edits. Training and evaluation use scene-disjoint splits;
details are given in Appendix~\ref{app:screening}.

\section{Experiments}
We evaluate TerrainForge at four levels: dynamics against CarSim; road geometry against real ZOD/LiDAR observations; intervention realization and multi-vehicle propagation in reconstructed Waymo scenes; and downstream scalability through a scene-disjoint intervention bank. Table~\ref{tab:setup} summarizes the primary experimental protocols. 

\begin{table}[t]
\centering
\caption{Primary experimental protocols.
Path replay fixes the recorded plan-view path while longitudinal and vertical
motion are re-integrated.}
\label{tab:setup}
\footnotesize
\setlength{\tabcolsep}{2.5pt}
\begin{tabular}{@{}>{\raggedright\arraybackslash}p{0.22\linewidth}>{\raggedright\arraybackslash}p{0.32\linewidth}>{\raggedright\arraybackslash}p{0.3\linewidth}>{\raggedright\arraybackslash}p{0.1\linewidth}@{}}
\toprule
Experiment & Control input fixed across arms & Vehicle motion & Camera \\
\midrule
CarSim (\S\ref{sec:carsim_validation}) & authored speed, steering, torque; identical in both simulators & 1 vehicle, re-solved & none \\
ZOD (\S\ref{sec:zod}) & measured speed and accelerations & half-car on the measured path & measured pose \\
Full chain, 109409 (\S\ref{sec:multivehicle}) & frozen inferred torque & ego + 17 moving vehicles, path replayed & dynamic \\
Multi-vehicle bank (\S\ref{sec:multivehicle}) & frozen inferred torque & 353 moving vehicles (ego included), path replayed & none \\
Video comparison (\S\ref{sec:generators}) & frames: none; response panel: speed tracking of recorded speed & frames: recorded schedule; panel: ego re-solved & kinematic \\
Surrogate bank (\S\ref{sec:screening}) & frozen inferred torque and steering & ego re-solved; target as recorded & none \\
\bottomrule
\end{tabular}
\end{table}
\subsection{Vehicle-model validation against CarSim}
\label{sec:carsim_validation}

We compare the coupled vehicle model with CarSim 2022.1 on five suites spanning low-speed, vertical-alignment, steering, and stress cases. Its parameters are fitted on development cases only. Table~\ref{tab:carsim_protocol} lists each suite's role in fitting, and its continuous errors; all comparisons use raw 5-ms traces without temporal alignment.

On four vertical alignments not used for fitting, speed RMSE is $0.004$--$0.007\,\mathrm{m/s}$ and pitch RMSE $0.051$--$0.061^\circ$ ($0.058^\circ$ crest, $0.051^\circ$ sag), and normal-load extrema differ by less than $0.002$ of static load. On the steered holdouts, lateral-position RMSE is $0.035$--$0.11$~m.

\begin{figure}[t]
\centering
\includegraphics[width=0.75\linewidth]{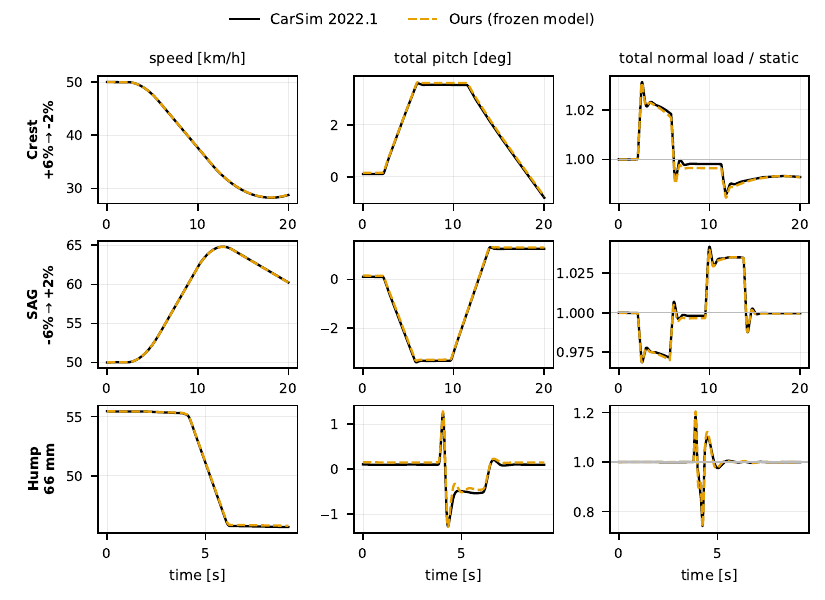}
\caption{
Representative CarSim validation on a crest, SAG, and speed hump.
Columns show longitudinal speed, body pitch, and total normal load normalized
by static load. Our frozen model closely tracks CarSim 2022.1 on the raw $5\,\mathrm{ms}$ trajectories; the hump case additionally
includes braking and steering.
}
\label{fig:carsim_vertical_events}
\end{figure}

\subsection{Real-road validation of the recovered road profile}
\label{sec:zod}

We validate the road profile used by the renderer and the vehicle model on
$96$ geometry-selected events from $13$ Zenseact Open Dataset (ZOD) recordings. The profile
$H(s)$ is recovered from smoothed OXTS altitude without fitting any ZOD-specific parameter,
and the measured speed and accelerations drive the inherited muSync-GS half-car with its
CarSim-identified parameters (Table~\ref{tab:setup}). On the held-out split, world-frame pitch RMSE is
$1.03^\circ/0.24^\circ/0.26^\circ$ for flat, kinematic, and half-car
predictions, respectively, showing that the improvement is primarily due to
the recovered road attitude. Against an independent LiDAR ground estimate,
the recovered profile gives $0.14^\circ$ error versus $0.45^\circ$ for a
flat-road assumption. Full protocol and per-type results are in Appendix~\ref{app:zod}.

\subsection{Scene realization and multi-vehicle counterfactual propagation}

\noindent\textbf{Geometry realization.}\label{sec:geometry_fidelity}
Across ten reconstructed Waymo scenes, 88 vertical curves and 30 humps
achieve median apex-depth and vertical-curvature errors of $0.17\%$ and
$0.05\%$, with a surface RMS of $1.2\,\mathrm{mm}$ (Appendix~\ref{app:geometry}).

\begin{figure}[t]
\centering
\begin{minipage}[c]{0.69\linewidth}
\includegraphics[width=\linewidth]{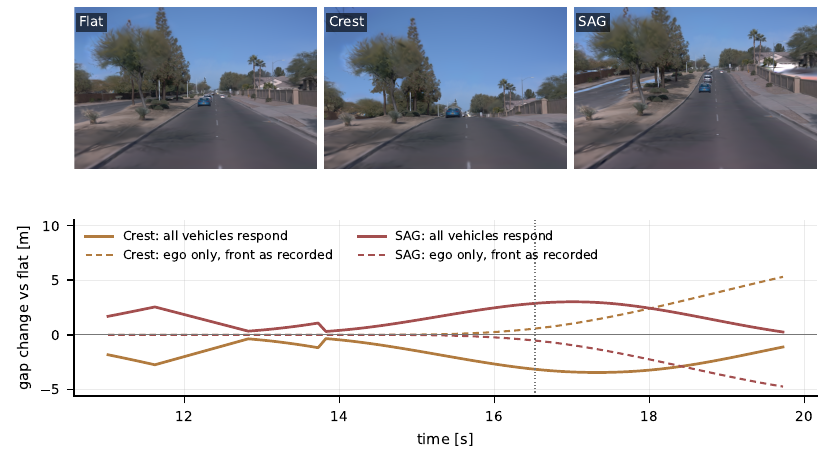}
\end{minipage}\hfill
\begin{minipage}[c]{0.29\linewidth}
\includegraphics[width=\linewidth]{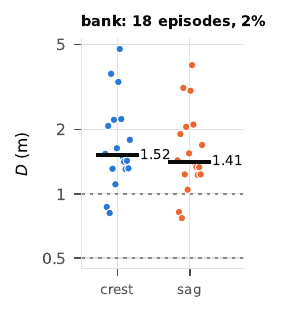}
\end{minipage}
\caption{
Multi-vehicle counterfactual propagation.
Left: Waymo 109409 under flat, crest, and sag edits, with the corresponding
ego--lead gap change for all-vehicle and ego-only propagation.
Right: peak disagreement $D$ over the 18 valid ego--lead episodes for the
primary $2\%$ edits; bars denote medians. Stress results are in Appendix~\ref{app:mv_bank}.}
\label{fig:full_chain}
\end{figure}

\noindent\textbf{Full-chain episode.}\label{sec:multivehicle}
On Waymo 109409, all $18$ moving vehicles replay their frozen inferred
controls on flat, crest, and sag roads. Nine traverse the edit and change
speed by up to $6.2$ and $5.6\,\mathrm{km/h}$, respectively, while the
others remain unchanged. Because the lead reaches the edit first,
re-simulating it reverses the sign of the ego--lead gap change relative to
replaying it as recorded (Figure~\ref{fig:full_chain}).

\noindent\textbf{Multi-vehicle generalization.}
We then evaluate the same propagation mechanism, without rendering, on
20 previously unused Waymo episodes comprising 353 valid moving-vehicle
rollouts, with the same control recovery and replayed plan-view paths. The edit
is a $120\,\mathrm{m}$ crest or sag centred on each ego route, at $2\%$ ($K=15$, the
primary arm) and $8\%$ ($K=3.75$, stress); episode selection, vehicle admission,
and the response threshold were frozen before any edited run
(Appendix~\ref{app:multivehicle}). Factual replay on the recorded road
has a median speed RMSE of $0.018\,\mathrm{m/s}$.

For the primary $2\%$ crest and sag interventions, $257/266$ ($96.6\%$) of
vehicles that traverse the edited region show a measurable physical response,
whereas all 87 non-traversing vehicles remain unchanged.

The multi-vehicle response also changes the interaction seen by the ego. In
each of 18 episodes with a usable ego--lead pair, the same edit is evaluated twice:
with every vehicle responding, and with only the ego responding while the lead keeps
its unedited rollout. Their ego--lead gap changes disagree by a median maximum of
$1.52\,\mathrm{m}$ (crest) and $1.41\,\mathrm{m}$ (sag), by more than $0.5\,\mathrm{m}$ in
every episode (Figure~\ref{fig:full_chain}; Appendix~\ref{app:mv_bank}). The minimum gap
changes less, by a median of $0.12$ and $0.10\,\mathrm{m}$, and is unchanged in $8$ of
$18$ episodes whose closest approach precedes the edit.

\noindent\textbf{Lateral coupling.}\label{sec:lateral}
On $177$ Waymo drives, $2\%$ edits change peak lateral acceleration by only
$0.004\,\mathrm{m/s^2}$ but the minimum wheel-load ratio by $0.022$
(Appendix~\ref{app:lateral}).
\begin{figure}[t]
\centering
\includegraphics[width=0.95\linewidth]{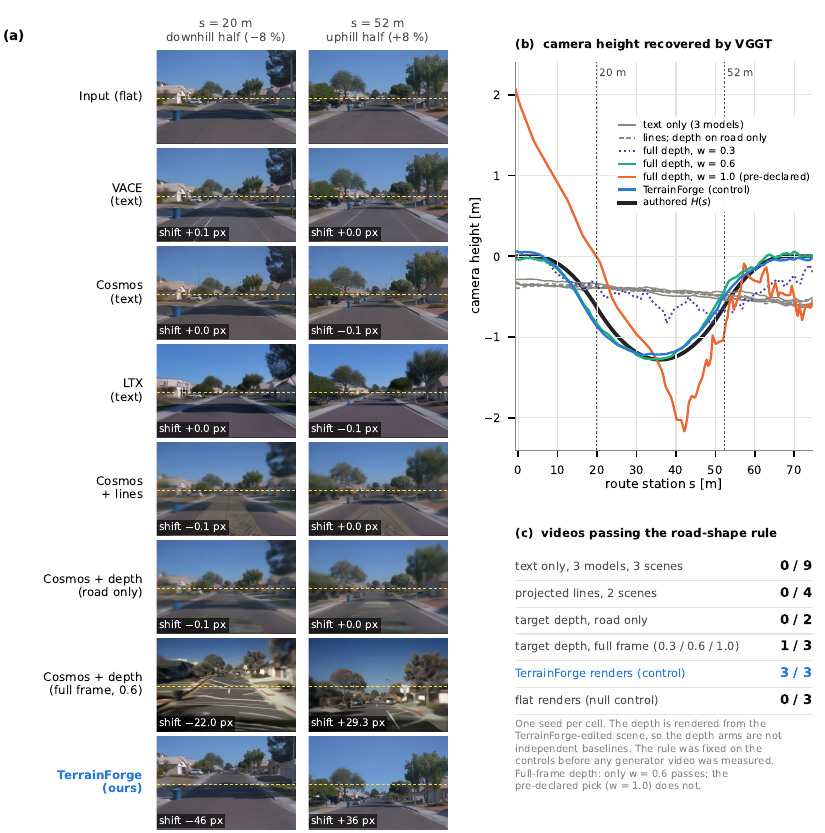}
\caption{
Comparison with video editing models on Waymo 150623.
(a) Frames at the steepest SAG stations; dashed lines mark the flat-road horizon.
(b) VGGT-recovered camera height versus the authored profile.
(c) Camera-trajectory consistency across conditioning modes. Only full-frame
depth reaches the target at one setting, while altering the recorded scene.}
\label{fig:generators_main}
\end{figure}
\subsection{Comparison with Video Editing Models}
\label{sec:generators}

We ask whether open video models, VACE~\citep{jiang2025vace,wan2025},
Cosmos-Transfer2.5~\citep{nvidia2025cosmosphysicalai}, and
LTX-Video~\citep{hacohen2024ltxvideo}, realize a prescribed sag or crest in a flat-road
clip while preserving it, given text alone, projected road lines, or target depth.
Generated pixels have no metric scale, so the criterion is \emph{camera-trajectory
consistency}: the camera path recovered by VGGT~\citep{wang2025vggt}, scaled to the
authored station span, must follow the target profile $H(s)$ in RMSE, correlation, and
peak-to-peak ratio, with thresholds fixed beforehand on ten controls
(Appendix~\ref{app:generators}). It is a necessary condition for realizing the edit; for
our own renders the road geometry is also checked directly on the deformed Gaussians (Section~\ref{sec:geometry_fidelity}).

Related multimodal work studies the complementary perception-side problem of
detecting traffic accidents from video~\citep{bian2025mllm}.

No text-only output is consistent with the target profile on any of three scenes ($0/9$):
VACE preserves the input and leaves the road unchanged, while Cosmos and LTX-Video
regenerate the scene without the requested motion. Projected lines ($0/4$) and depth on the
road only ($0/2$) do not pass either. Depth over the whole frame, rendered from the
TerrainForge-edited scene and so already carrying the authored geometry, passes at one of
three control weights on Cosmos (RMSE $0.10\,\mathrm{m}$, correlation $0.98$), but that output
changes the scene outside the road by $44.6/255$ against $4.5$--$15.5/255$ for text-only
outputs: geometric control is bought with fidelity to the recording.
\subsection{Screening candidate road edits}
\label{sec:screening}

We train the surrogate of Section~\ref{sec:surrogate} on a scene-disjoint
Waymo bank of $15{,}758$ road edits ($11{,}834$ sag and $3{,}924$ hump)
over $983$ braking episodes. Only the ego is re-simulated, while the
vehicle, pedestrian, or cyclist target follows its recorded trajectory.
The surrogate predicts terminal-gap change $\Delta G$, minimum-clearance
change $\Delta d_{\min}$, and whether an edit materially changes the outcome.
Full construction and protocol are given in Appendix~\ref{app:screening}.

\begin{table}[t]
\centering
\caption{Terminal-gap prediction on held-out scenes.
MAE is in meters; ratio is relative to predicting no change; $\rho$ is
Spearman rank correlation. Brackets denote $95\%$ scene-bootstrap intervals.}
\label{tab:screening}
\small
\setlength{\tabcolsep}{5pt}
\begin{tabular}{lcccc}
\toprule
Edit & No-change MAE & Ours MAE & Ratio & $\rho$ \\
\midrule
SAG  & $0.451$ & $0.270$ & $0.60$ [$0.41$, $0.76$]
     & $0.89$ [$0.78$, $0.96$] \\
Hump & $0.730$ & $0.569$ & $0.78$ [$0.44$, $0.89$]
     & $0.81$ [$0.72$, $0.89$] \\
\bottomrule
\end{tabular}
\end{table}

On held-out scenes, $\Delta G$ is predicted with MAE
$0.27\,\mathrm{m}$ for SAGs and $0.57\,\mathrm{m}$ for humps, reducing
error by $40\%$ and $22\%$ relative to predicting no change, with
$\rho=0.89$ and $0.81$, respectively. Minimum-clearance change is not
predicted better than the no-change baseline
(Table~\ref{tab:screening_full}). For consequential-edit detection, the
classifier reaches AP $0.85$ [$0.78$, $0.90$], compared with $0.20$ for
random ranking and $0.35$ using edit geometry alone. The classification
label is direction-agnostic and therefore includes both beneficial and
harmful changes.

\section{Conclusion}
TerrainForge constructs road counterfactuals in reconstructed driving
episodes by applying a shared road intervention to vehicle dynamics, scene
geometry, and camera motion. Validation against CarSim, ZOD/LiDAR, and
reconstructed Waymo scenes supports the physical and geometric components,
while multi-vehicle experiments show that surrounding-vehicle response can
change estimated ego--lead interaction. A scene-disjoint surrogate further
predicts terminal-gap changes for inexpensive screening of candidate edits.
\bibliographystyle{plainnat}
\bibliography{references}

\appendix
\section{Additional method details}
\label{app:method}

\paragraph{Road-geometry increments.}
\label{app:road_profiles}
Continuous alignments use tangent grades joined by parabolic vertical curves. For a curve of length $L$ changing grade from $g_{1}$ to $g_{2}$, with $\xi$ measured from its beginning,
\begin{equation}
\begin{split}
H(\xi)&=H_{B}+g_{1}\xi+\frac{g_{2}-g_{1}}{2L}\xi^{2},\\
H'(\xi)&=g_{1}+\frac{g_{2}-g_{1}}{L}\xi,
\qquad H''(\xi)=\frac{g_{2}-g_{1}}{L}.
\end{split}
\label{eq:vertical_curve}
\end{equation}
Road pitch is $\alpha(s)=\arctan H'(s)$ and vertical curvature is $\kappa_v(s)=H''(s)/(1+H'(s)^2)^{3/2}$. A compact feature of height $A$, length $L_h$, and center $s_c$ is a raised cosine, $H_h(s)=\tfrac{A}{2}\left[1+\cos\!\big(2\pi(s-s_c)/L_h\big)\right]$ for $|s-s_c|\leq L_h/2$ and zero outside its footprint, so its zero endpoint height and grade avoid discontinuous wheel excitation and visible seams. The same authored profile is materialized for dynamics, Gaussian deformation, camera motion, and geometry verification.

\paragraph{Friction field.}
Friction is the third component of $\mathcal{R}_{\eta}$ and lives in the same road frame. A surface has a base coefficient $\mu_{0}$ and optional patches $\Omega_{p}$ that override it,
\begin{equation}
\mu_{\eta}(x,y)=
\begin{cases}
\mu_{p}, & (x,y)\in\Omega_{p},\\
\mu_{0}, & \text{otherwise},
\end{cases}
\label{eq:friction_field}
\end{equation}
where each patch takes the longitudinal and lateral footprint of a compact feature and later patches take precedence. Here $\mu$ is the peak tire--road friction coefficient of the tire model. It is an authored variable: $\mu_{0}$ is declared for each study, factual grounding (Section~\ref{sec:grounding}) is run on the source surface, and the frozen controls are then replayed on $\mu_{\eta}$. The effect of a friction edit is therefore the change in outcome of identical controls.

\paragraph{Coupled vehicle model.}
The coupled state is
\begin{equation}
\begin{split}
\mathbf{x}=[&X,Y,\psi,v_x,v_y,r,z_s,\dot z_s,\theta,\dot\theta,\phi,\dot\phi,\\
&z_u^{1:4},\dot z_u^{1:4},\kappa^{1:4},\alpha_t^{1:4},\omega^{1:4}],
\end{split}
\label{eq:state}
\end{equation}
which contains 32 states spanning planar motion, heave, pitch, roll, four unsprung masses, dynamic longitudinal and lateral slip, and wheel rotation. At every integration substep, current body states determine the wheel positions. Each wheel queries~\eqref{eq:road_field} for contact height, surface gradient, and friction $\mu_{i}=\mu_{\eta}(x_{i},y_{i})$; suspension and tire compression yield $F_{z,i}$, and dynamic TMeasy responses produce $F_{x,i}$ and $F_{y,i}$. The planar equations are written in body-frame form; road-gradient gravity enters through the contact-frame force projection $F_{x,i}^{b},F_{y,i}^{b}$,
\begin{equation}
\begin{split}
\dot X &=v_x\cos\psi-v_y\sin\psi,\quad
\dot Y=v_x\sin\psi+v_y\cos\psi,\\
\dot v_x&=\textstyle\sum_i F_{x,i}^{b}/m+r v_y,\quad
\dot v_y=\textstyle\sum_i F_{y,i}^{b}/m-r v_x,\\
\dot r&=\textstyle\sum_i(x_iF_{y,i}^{b}-y_iF_{x,i}^{b}+M_{z,i})/I_z.
\end{split}
\label{eq:planar}
\end{equation}
Here $r$ is yaw rate, $I_z$ is yaw inertia, and $M_{z,i}$ is the tire aligning moment. Slip states follow relaxation dynamics, e.g., $\dot\kappa_i=(\kappa_i^{\rm kin}-\kappa_i)v_{{\rm tr},i}/L_{x,i}$, while wheel speed obeys $I_w\dot\omega_i=T_i-RF_{x,i}+M_{y,i}$, with $L_{x,i}$ the longitudinal relaxation length, $v_{{\rm tr},i}$ the contact-patch transport speed, $R$ the effective wheel radius, and $M_{y,i}$ the wheel rotational moment. Longitudinal and lateral forces share the same capacity and are projected when necessary to satisfy
\begin{equation}
\sqrt{F_{x,i}^{2}+F_{y,i}^{2}}\leq \mu_iF_{z,i}.
\label{eq:friction_circle}
\end{equation}
Thus road elevation changes contact and load, load changes tire capacity, tire forces change world motion and body attitude, and updated wheel positions query the road again at the next substep.

\paragraph{Control recovery in the single-vehicle banks.} The lateral and obstacle banks re-solve planar motion and use a feedforward inverse of the model instead of iterative learning. Body-frame speed is smoothed by $\mathrm{SG}_{11,3}$ and differentiated to $a_{x}$, and yaw rate $r$ is the derivative of $\mathrm{SG}_{11,2}$-smoothed heading. With $F^{0}_{z,i}$ the static wheel load and $F^{\rm g}_{x,j}$ the body-frame component of gravity along the source-road gradient at wheel $j$,
\begin{equation}
T_{i}=R\,\frac{F^{0}_{z,i}}{\sum_{j}F^{0}_{z,j}}\Big[m\,(a_{x}-r\,v_{y})-\textstyle\sum_{j}F^{\rm g}_{x,j}\Big]+\frac{I_{w}\,a_{x}}{R},
\label{eq:feedforward_torque}
\end{equation}
followed by a slew limit of $3000\,\mathrm{N\,m\,s^{-1}}$ on a $1.25\,\mathrm{ms}$ actuator clock. Both front wheels receive $\delta_{f}=\arctan\!\big(L\,r/\max(v_{x},0.5\,\mathrm{m/s})\big)$ with $L=2.7\,\mathrm{m}$; rear steering is zero. The obstacle target of the screening bank is replayed as recorded.

\paragraph{Scene deformation.}\label{app:scene_deformation}
Gaussian primitives are projected into the $(s,d,h)$ frame and assigned a surface-support category from semantics and geometric support. Road and sidewalk surfels follow the displacement and rotate to the new surface normal; roadside terrain follows elevation while remaining gravity aligned; supported objects move rigidly according to their footing; unrelated or out-of-corridor geometry remains fixed. For primitive $i$,
\begin{equation}
\mathbf{p}'_i=\mathbf{p}_i+w_i\Delta H_{\eta}(s_i)\mathbf{e}_z,
\label{eq:gaussian_deform}
\end{equation}
where $w_i$ encodes lateral influence and attachment. Surface-resting rigid objects are additionally rotated about their footing to avoid penetration or floating. The rollout supplies the vehicle pose $T_v^{\eta}(t)$, and with a fixed vehicle-to-camera calibration the counterfactual camera is $T_c^{\eta}(t)=T_v^{\eta}(t)T_{v\rightarrow c}$.

\paragraph{Traffic participants as rigid Gaussian objects.}
Vehicles other than the ego respond differently on the edited road, so they cannot remain recorded pixels. Re-estimating each vehicle's ground contact and appearance from two-dimensional masks in every frame produces floating, popping, and flicker, because every frame repeats the estimate. We instead reconstruct each tracked vehicle that is observed by at least one camera once, in its own box frame, and move it as a rigid body. Vehicle Gaussians that remain in the background without a track follow the surface-resting rule above.

\paragraph{Reconstruction.}
Vehicle $a$ has a labeled box pose $B^{a}_{k}$ and size at frame $k$. Its model $\mathcal{G}^{a}$ is a set of two-dimensional Gaussian surfels in the box frame, initialized from the LiDAR returns of all sensors inside the box, expressed in that frame and accumulated over time. Ground is removed frame by frame from a ring of returns just outside the footprint; a fixed height cut would slice the tires and leave the vehicle floating by the cut height. Supervision comes from five cameras. Each vehicle's own projected three-dimensional box prompts a SAM~2 mask, confined to the box hull, with the nearer box winning on overlap, so a mask carries its track identity by construction. Four choices make sparse, one-sided views sufficient. Geometry is mirror-symmetric about the longitudinal plane, so a vehicle seen from one side still has a whole body, while color is free on each side because the sun lights them differently. The front camera fixes the reference color and every other camera learns an affine exposure correction. Opacity is supervised only where it is known: one inside the vehicle's own mask, zero inside its projected box hull but outside every vehicle mask, and free where another vehicle's mask lies, since the vehicle may be behind it. Finally, centimetre-level label noise averaged over hundreds of views blurs a vehicle, so each frame has a rigid correction $\Delta_{k}\in SE(3)$ shared by the cameras of that frame, active after a warm-up, and the vehicle is rendered at $B^{a}_{k}\Delta_{k}$ during training. The objective is
\begin{equation}
\mathcal{L}_{a}=\lambda_{1}\mathcal{L}_{1}+\lambda_{2}\mathcal{L}_{\rm SSIM}+\lambda_{3}\mathcal{L}_{\alpha}+\mathcal{L}_{\rm reg},
\label{eq:object_loss}
\end{equation}
with the photometric terms taken inside the vehicle's mask, $\mathcal{L}_{\alpha}$ a binary cross-entropy on the supervised opacity region, and $\mathcal{L}_{\rm reg}$ collecting a weak tie between the two side colors, the deviation of the exposure corrections from identity, an anisotropy penalty on surfels, and the magnitude of $\Delta_{k}$. Weights are fixed and shared by all vehicles. The corrections only sharpen the model and are not used afterwards. A vehicle that never reaches a minimum mask size ($120$ pixels) in any camera is not reconstructed and is not rendered.

\paragraph{Motion and placement.}
Each vehicle has two poses, and they serve different purposes. The \emph{metric} counterfactual pose is the vehicle's recorded path advanced by the difference between its edited and baseline rollouts (Section~\ref{sec:grounding}), so a zero edit reproduces the recording:
\begin{equation}
\begin{aligned}
\mathbf{p}^{\rm cf}_{a}(t)&=\boldsymbol{\gamma}_{a}\!\big(q_{a}(t)+\Delta q_{a}(t)\big)+\big[\Delta H(s)+\Delta h_{a}(t)\big]\hat{\mathbf{e}}_{z},\\
R^{\rm cf}_{a}(t)&=S\big(\Delta H'(s)\big)\,R^{\rm rec}_{a}(t)\,\mathcal{B}\big(\Delta\theta_{a}(t),\Delta\phi_{a}(t)\big),
\end{aligned}
\label{eq:metric_pose}
\end{equation}
where $\boldsymbol{\gamma}_{a}$ is the recorded path parameterized by its own arc length $q_{a}$, $\Delta q_{a}$ and the heave $\Delta h_{a}$ are edited-minus-baseline rollout differences, $S$ rotates the vertical onto the edited road normal, and $\mathcal{B}$ applies the rollout's body pitch and roll, both taken \emph{relative to the road} and as edited-minus-baseline differences, so that $\mathcal{B}$ carries suspension response only. Minimum signed clearance, collision, and the terminal gap (\emph{Safety metrics} below) are computed on $(\mathbf{p}^{\rm cf}_{a},R^{\rm cf}_{a})$ in the metric Waymo frame and on nothing downstream of it. Before a counterfactual onset time (in our scene, the ego's launch) every vehicle keeps its recorded pose.

The \emph{displayed} pose is obtained from the metric one by a chain of corrections that keeps each vehicle on the pixels beneath its wheels:
\begin{equation}
\mathbf{p}^{\rm cf}_{a}
\xrightarrow{\;-\Delta H,\ S^{-1}\;}\bar{\mathbf{p}}_{a}
\xrightarrow{\;A_{k}\;}\mathbf{x}^{(0)}_{a,k}
\xrightarrow{\;\text{Eq.~\ref{eq:contact_height}}\;}\mathbf{x}^{(1)}_{a,k}
\xrightarrow{\;\text{Eq.~\ref{eq:depth_factor}}\;}\mathbf{x}^{(2)}_{a,k}
\xrightarrow{\;\text{Eq.~\ref{eq:edit_lift}}\;}\mathbf{x}^{(3)}_{a,k}
\xrightarrow{\;\text{temporal smoothing}\;}\mathbf{x}^{\rm disp}_{a,k}.
\label{eq:display_chain}
\end{equation}
The authored lift and tilt are first removed, since they are re-applied from the drawn road below. $A_{k}$ maps the pose into the reconstruction frame through the calibrated camera that renders frame $k$, not through one global similarity, because structure-from-motion warps the world along a curved drive and a single transform places a vehicle meters from where the images put it. Contact height is then taken from the unedited scene's road where it is reconstructed:
\begin{equation}
z_{a}=w_{a}\,\hat z_{\rm flat}+(1-w_{a})\,z_{a}^{\rm label},\qquad
w_{a}=\frac{1}{4}\sum_{i=1}^{4}\operatorname{clip}\!\left(\frac{n_{i}-n_{0}}{n_{1}-n_{0}},0,1\right),
\label{eq:contact_height}
\end{equation}
where $\hat z_{\rm flat}$ is a smoothed median-height field over the road Gaussians of the unedited scene, read at the four wheel contacts and fitted by a plane whose value at the vehicle center is used, $z_{a}^{\rm label}$ is the height implied by the labeled pose, and $n_{i}$ counts road Gaussians within $1.5\,\mathrm{m}$ of contact $i$ ($n_{0}=5$, $n_{1}=30$). Where the reconstruction has road under the wheels the vehicle sits on it; where it has none, such as an intersection that vehicles covered in every training image, the labeled height is used. One weight is shared by all four wheels, because the two surfaces disagree by $0.3$--$0.7\,\mathrm{m}$ at the intersection and blending them wheel by wheel invents a slope across the vehicle. For the same reason the road field sets height only, never attitude: the displayed attitude before the edit is re-applied is $R^{\rm rec}_{a}\mathcal{B}(\Delta\theta_{a},\Delta\phi_{a})$.

\paragraph{Depth consistency.}
The reconstructed road can also lie at a different depth from the labeled one. To rest a vehicle on the surface that is drawn, it is scaled about the camera center $\mathbf{c}_{k}$,
\begin{equation}
\rho_{a,k}=\operatorname*{median}_{i}\frac{D_{k}(\mathbf{u}_{i})}{z_{i}},\qquad
\mathbf{x}\mapsto\mathbf{c}_{k}+\rho_{a,k}\left(\mathbf{x}-\mathbf{c}_{k}\right),
\label{eq:depth_factor}
\end{equation}
where $D_{k}$ is the depth rendered for the unedited scene and $\mathbf{u}_{i}$, $z_{i}$ are the pixel and camera depth of wheel contact $i$. Since $\pi_{k}\big(\mathbf{c}_{k}+\rho(\mathbf{x}-\mathbf{c}_{k})\big)=\pi_{k}(\mathbf{x})$ for the projection $\pi_{k}$ of camera $k$, and the vehicle's extent scales with its distance, its pixel position and size are unchanged and only its depth moves. What the scaling can change is which surface wins the depth test, between vehicles or between a vehicle and the road. The factor is measured once on the unedited render, smoothed in time, and reused in every case: on an edited render a hidden vehicle's wheels would read the crest itself and the vehicle would be pulled in front of it. Where the drawn road is a thick layer of semi-transparent floaters ($\rho_{a,k}<0.8$), the vehicle is moved up to a further $20\%$ toward the camera, again along its ray, so that it does not blend into that layer. On 109409, over the $582$ vehicle-frames in which a vehicle is on screen, the applied factor has $10$th/$50$th/$90$th percentiles $0.89$/$0.96$/$0.98$ after the onset, when vehicles are on the driven road.

The edit is then added once, as the change the edit made to the drawn road under the scaled contact $\mathbf{x}_{c}$:
\begin{equation}
\begin{aligned}
\delta_{a,k}&=\tau\,\big[\hat z_{\rm case}(\mathbf{x}_{c})-z(\mathbf{x}_{c})\big]+(1-\tau)\,\big[\hat z_{\rm case}(\mathbf{x}_{c})-\hat z_{\rm flat}(\mathbf{x}_{c})\big],\qquad
\tau=\operatorname{clip}\!\left(\frac{\rho_{a,k}-0.6}{0.3},0,1\right),\\
R&\mapsto T\big(\nabla\hat z_{\rm case}-\nabla\hat z_{\rm flat}\big)\,R,
\end{aligned}
\label{eq:edit_lift}
\end{equation}
where $\hat z_{\rm case}$ is the road field of the edited scene and $T$ tilts the vertical onto the normal of the slope \emph{difference}. Where the road is properly reconstructed ($\tau=1$) the vehicle is set onto the drawn edited road; on a painted floater layer ($\tau=0$) it receives only the change of the drawn road, which is zero where the edit does not reach. The tilt vanishes on the unedited case and before the onset, where the height term only re-seats the depth-scaled vehicle on the drawn road. Finally, the vertical offset of $\mathbf{x}^{(3)}$ from the depth-scaled labeled position is Savitzky--Golay filtered in time ($21$ frames, order $2$), since the road field is interpolated on most of its cells and otherwise makes vehicles bob as they cross them.

The displayed attitude is therefore $T\,R^{\rm rec}_{a}\,\mathcal{B}(\Delta\theta_{a},\Delta\phi_{a})$, the display counterpart of $R^{\rm cf}_{a}$ in Eq.~\ref{eq:metric_pose}: the grade enters once, through $T$, and $\mathcal{B}$ is road-relative, so the grade is not counted twice. The chain of Eq.~\ref{eq:display_chain} moves a vehicle only along its camera ray (Eq.~\ref{eq:depth_factor}) and vertically (Eqs.~\ref{eq:contact_height} and~\ref{eq:edit_lift}), so the rendered image of a vehicle departs from the projection of its metric box only through the vertical corrections. We measure this on 109409 by projecting the metric box and the displayed box through the camera that renders each frame. After the onset, on the crest and SAG cases, the bottom edge of the displayed vehicle lies a median $0.15$--$0.16\,\mathrm{m}$ ($95$th percentile $0.40$--$0.43\,\mathrm{m}$, or $4$--$5$ and $12$--$13$ pixels) from that of the metric box at the vehicle's range, the same as on the unedited case ($0.17$ and $0.34\,\mathrm{m}$), so the deviation is the reconstruction's road offset rather than an effect of the edit; horizontal offsets stay below $1.6$ pixels. The displayed body pitch differs from the metric pitch by a median $0.2^{\circ}$ ($95$th percentile $0.7^{\circ}$) and is exactly zero on the unedited case; regressed on the pitch the edit itself adds, this residual has slope $-0.02$, where a grade counted twice would give $1$ and a grade omitted $-1$. Clearances and distances are measured on the metric poses.

\paragraph{Track continuity and rendering.}
Some tracks begin or end while the vehicle is still in view, which would make it pop in or out. Such a track is extended at the constant velocity and yaw rate of its first or last second, and only while the extended box projects into the recorded camera with a real area and the recorded frame shows vehicle pixels inside it ($\geq 35\%$). The frames are decided once on the recorded track and reused by every case. Footprints of tracked vehicles along their recorded trajectories are removed from the background Gaussians; the surfels of every vehicle are transformed by their per-frame poses and rasterized together with the deformed background from the rollout camera in one pass. Occlusion by a crest or by another vehicle is then the depth test rather than a hand-written rule.

\paragraph{Weather and surface-condition rendering.}\label{app:weather}
Friction changes vehicle response but not, by itself, the image. Weather is therefore an optional appearance layer applied to finished geometry and motion, and it is authored beside $\mu_{\eta}$ rather than derived from it: the intervention specification carries a visual condition (rain rate, snow level) and a friction coefficient as separate variables. We orchestrate the weather-editing model and precipitation compositors of the muSync-GS pipeline~\citep{musyncgs2026} without modifying them. What we add is one rule: every buffer these components consume is rendered from the edited scene and the rollout's camera poses, never reused from the unedited drive.

\paragraph{Rain.}
The static appearance edit runs on frames re-rendered from the edited scene at the rollout's poses, with segmentation recomputed on those frames, since a grade changes the relative position of buildings, trees, and road. A thin water film is added by screen-space reflection from road-mask, depth, and normal buffers rendered from the same scene and poses. Precipitation particles are rendered in a world-vertical frame at the telemetry-bound camera pose, so a camera pitched by a grade does not tilt the rain. Camera pitch and suspension heave are applied once, by the final compositor, and are not also present in the pose sequence handed to it.

\paragraph{Snow.}
Snow is a removable layer of thin surfels placed a fixed metric distance ($30\,\mathrm{mm}$) above every top-facing road and sidewalk surfel of the final Gaussian scene, in coherent patches so that partial cover forms contiguous bare and snowy regions rather than noise. No station coordinate or centerline is used, so the rule holds for an edited profile and for any trajectory shape. Its appearance enters as the residual between a render of the snowy scene and a render of the same scene without the layer, from the same poses, and is composited with the background edit and rendered snow particles.

\begin{figure}[t]
\centering
\includegraphics[width=\linewidth]{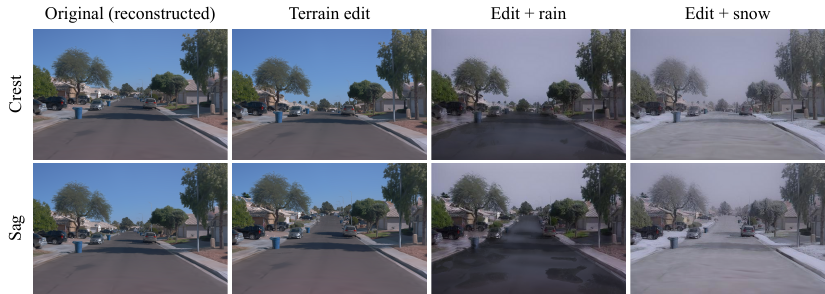}
\caption{Rain and snow on vertical-alignment edits of Waymo scene 150623. Rows: crest and
sag ($\pm6\%$, $K=7.5\,\mathrm{m}/\%$). Columns: reconstruction, terrain edit, edit with
rain, and edit with snow, all at the same frame.}
\label{fig:weather_quality}
\end{figure}

Figure~\ref{fig:weather_quality} shows both layers on a crest and a sag; the quantitative evaluation in this paper concerns geometry and dynamics.

\paragraph{Safety metrics.}
All safety quantities are computed from simulated states in the metric world frame; the display corrections of Eq.~\ref{eq:display_chain}, including the depth factor of Section~\ref{sec:traffic_objects}, affect rendering only. Each vehicle or target $a$ occupies a planar oriented box $B^{a}(t)\subset\mathbb{R}^{2}$ given by its labeled length and width and its position and yaw at time $t$; height is not used. The signed clearance of the ego $E$ and a target $T$ is
\begin{equation}
c(t)=\begin{cases}
\operatorname{dist}\!\big(B^{E}(t),B^{T}(t)\big), & B^{E}(t)\cap B^{T}(t)=\emptyset,\\[2pt]
-\min_{\mathbf{n}\in\mathcal{N}(t)}\mathrm{ov}_{\mathbf{n}}(t), & \text{otherwise},
\end{cases}
\label{eq:signed_clearance}
\end{equation}
where $\mathcal{N}(t)$ are the four edge normals of the two boxes and $\mathrm{ov}_{\mathbf{n}}$ is the overlap of their projections onto $\mathbf{n}$, so a negative value is the penetration depth along the separating axis of least overlap. Contact is tested at every $5\,$ms step and the first contact time is $t_{c}=\min\{t:c(t)\le 0\}$ ($t_{c}=\infty$ if none). The minimum clearance is $d_{\min}=\min_{t}c(t)$ over the valid horizon, with $c$ evaluated every $20\,$ms and at the steps adjacent to any contact. It is evaluated for the ego and its designated target.

The terminal gap is measured along the ego route, with $s$ the route station. Let $e^{a}(t)=\tfrac12\big(L^{a}|\cos\Delta\psi^{a}|+W^{a}|\sin\Delta\psi^{a}|\big)$ be the half-extent of box $a$ along the route tangent, where $\Delta\psi^{a}$ is its heading relative to the route, and
\begin{equation}
g(t)=\big(s^{T}(t)-e^{T}(t)\big)-\big(s^{E}(t)+e^{E}(t)\big).
\label{eq:route_gap}
\end{equation}
The ego is stopped at $t_{s}$, the first time from which its speed stays at or below $0.1\,\mathrm{m/s}$ for $0.5\,$s. The terminal gap is $G=g(t^{*})$ with $t^{*}=t_{s}$ if the ego stops within the valid horizon and $t^{*}=t_{\rm end}$ otherwise; the valid horizon is identical for the two arms of a paired comparison. We report $\Delta G_{\eta}=G(\mathcal{R}_{\eta})-G(\mathcal{R}_{0})$. $G$ ignores lateral offset, so it is negative once the ego has passed a target beside its path; only its change is used. When both arms stop, $\Delta G$ is a stopping-margin change; otherwise it measures the change in terminal along-route gap over the common horizon.

Post-contact response is not modeled, and after $t_{c}$ the boxes pass through each other. A rollout with $t_{c}<\infty$ is therefore labeled a collision, which alone makes the edit effective, and is reported with its contact time and relative speed. Its continuous quantities are kept in the regression targets as simulated, and $d_{\min}$ then includes the penetration depth.

In the multi-vehicle experiments, interaction is summarized by the ego--lead gap: the difference of the two vehicles' along-path distances, whose change under an edit is the change of the lead's path distance minus the change of the ego's. All vehicle pairs are additionally checked for box overlap as a validity condition.

\paragraph{Evaluated objects.}
Table~\ref{tab:object_scope} lists, for each experiment, which objects are re-simulated, rendered, and scored. Dynamic rollouts are defined for tracked vehicles. In the obstacle bank, pedestrian and cyclist targets follow their recorded trajectories. Rendering uses reconstructed tracked vehicles when available; untracked content remains part of the static background.

\begin{table}[h]
\centering
\caption{Object scope of each experiment. ``Replayed'' objects follow their recorded trajectories in both arms of a comparison.}
\label{tab:object_scope}
\small
\begin{tabular}{@{}>{\raggedright\arraybackslash}p{0.17\linewidth}>{\raggedright\arraybackslash}p{0.33\linewidth}>{\raggedright\arraybackslash}p{0.2\linewidth}>{\raggedright\arraybackslash}p{0.24\linewidth}@{}}
\toprule
Experiment & Re-simulated & Rendered as objects & Safety quantity \\
\midrule
Waymo 109409 (video) & ego and $17$ moving vehicles; $5$ parked vehicles replayed & $19$ of $22$ tracked vehicles$^{\dagger}$ & ego--lead gap; all-pair overlap check \\
Multi-vehicle bank ($20$ episodes) & $353$ moving vehicles (ego included); parked vehicles, pedestrians, cyclists not modeled & none & ego--lead gap \\
Obstacle bank & ego only; target replayed & none & ego--target $d_{\min}$, $\Delta G$, collision \\
Lateral bank & ego only & none & peak lateral and load quantities \\
\bottomrule
\end{tabular}

\vspace{2pt}\raggedright{\footnotesize $^{\dagger}$Three vehicles never exceed $120\,$px in any camera (two travel directly behind the ego) and are not reconstructed.}
\end{table}

\begin{table}[h]
\centering
\caption{Settings of the secondary experiments, in the format of Table~\ref{tab:setup}.}
\label{tab:setup_extra}
\footnotesize
\setlength{\tabcolsep}{2.5pt}
\begin{tabular}{@{}>{\raggedright\arraybackslash}p{0.15\linewidth}>{\raggedright\arraybackslash}p{0.25\linewidth}>{\raggedright\arraybackslash}p{0.21\linewidth}>{\raggedright\arraybackslash}p{0.11\linewidth}>{\raggedright\arraybackslash}p{0.19\linewidth}@{}}
\toprule
Experiment & Control input fixed across arms & Vehicle motion & Camera & Checks \\
\midrule
Scene 22 (App.~\ref{app:multivehicle}) & prescribed: $36\,$km/h, $1000\,$N\,m brake from $5.34\,$s & ego re-solved; lead static & dynamic & one braking mechanism \\
Lateral edit (\S\ref{sec:lateral}) & frozen inferred torque and steering & ego, re-solved & none & vertical$\rightarrow$lateral coupling \\
Screening, lateral (App.~\ref{app:screening}) & frozen inferred torque and steering; $\mu$ varied & ego, re-solved & none & labels for the surrogate \\
\bottomrule
\end{tabular}
\end{table}

\paragraph{Intervention selection.}
Collision, low-clearance, wheel-contact, friction-utilization, road-support, and numerical-convergence checks are evaluated over the complete supported horizon. Given a discrete action set $\mathcal{A}$, intervention design is
\begin{equation}
\eta^{*}=\arg\min_{\eta\in\mathcal{A}} C(\eta)
\quad\mathrm{s.t.}\quad d_{\min}(\eta)\geq d_{\rm safe},\;V(\eta)=1,
\label{eq:intervention}
\end{equation}
where $C$ is infrastructure cost and $V$ denotes physical, geometric, and numerical validity. We retain ineffective, harmful, and unsupported candidates. A learned response or ranking model, when used, proposes actions only; final labels are always obtained from the full coupled rollout.

\section{Additional CarSim results}
\label{app:carsim}

\paragraph{Validation suites.} Table~\ref{tab:carsim_protocol} lists every suite with the model that produced it, its role, and its continuous errors. Roles: \emph{F}, used to fit parameters; \emph{R}, regression on development cases;
\emph{E}, blind evaluation cases not inspected until the model parameters were frozen.
\begin{table}[t]
\centering
\caption{CarSim comparison suites. Errors are RMSE ranges over the cases of a suite on raw
$5\,$ms traces; a case passes when every pre-declared accuracy, integrity, and
numerical-convergence gate is met. All suites score the coupled vehicle model except grade-zero,
which is inherited from muSync-GS and scores its half-car.}
\label{tab:carsim_protocol}
\scriptsize
\setlength{\tabcolsep}{3pt}
\begin{tabular}{@{}l r c p{5.6cm} l@{}}
\toprule
Suite & Cases & Role & Continuous error (RMSE) & Passed \\
\midrule
Grade-zero regression & 19 & R & event pitch mean $0.035^\circ$, max $0.052^\circ$; speed $0.023\,$m/s & 19/19 \\
Development & 5 & F & lateral position $0.030$--$0.157\,$m; pitch $0.041$--$0.068^\circ$ & 5/5 \\
Steered holdout & 3 & E & lateral position $0.035$--$0.110\,$m; pitch $0.046$--$0.067^\circ$; speed $0.010$--$0.020\,$m/s & 3/3 \\
Low speed & 4 & E & speed $0.002$--$0.005\,$m/s; pitch $0.050$--$0.066^\circ$; lateral position $\le0.28\,$m & 4/4 \\
Vertical alignment & 4 & E & speed $0.004$--$0.007\,$m/s; pitch $0.051$--$0.061^\circ$; heave $0.7$--$2.0\,$mm & 4/4 \\
Scene-supported SAG & 4 & E & speed $\le0.001\,$m/s; pitch $0.064$--$0.067^\circ$; lateral position $2.5$--$10.4\,$mm & 4/4 \\
SAG stress & 6 & E & passing cases: pitch $0.063$--$0.072^\circ$; failed case: speed $0.157\,$m/s (limit $0.12$) & 5/6 \\
\bottomrule
\end{tabular}
\end{table}

\paragraph{Extended description of the main-text results.}
The grade-zero benchmark is reported for continuity with muSync-GS: event-window
pitch RMSE of its half-car averages $0.035^\circ$ (maximum $0.052^\circ$ against a
$0.055^\circ$ threshold) and $0.025^\circ$ over the full horizon. Equations and thresholds were never changed. Suites scored before the final front lateral
tire-force scale were re-scored with it against the unchanged CarSim traces, and no pass or fail
verdict changed; for the low-speed suite, whose traces could not be re-scored, the difference
between the rollouts under the earlier and final scale is too small to change any verdict.

\paragraph{Vertical alignment.}
To test the road-input path directly, we use four profiles that were not used to
identify any parameter: $+6\%$ and $-6\%$ sustained grades, a crest
($+6\%\to-2\%$), and a SAG ($-6\%\to+2\%$), the last two with a $75\,\mathrm{m}$
vertical curve. Each case coasts from $50\,\mathrm{km/h}$ with no braking or
steering, so the road alone moves the vehicle. All four pass every gate.
Figure~\ref{fig:carsim_vertical_events} shows the crest and SAG, together with the
$66\,\mathrm{mm}$ speed-hump holdout that includes braking and steering.

On the crest, gravity alone reduces speed from $50$ to $28.3\,\mathrm{km/h}$ in
both simulators; on the SAG it raises speed to $64.8\,\mathrm{km/h}$. Total
normal load stays within $0.986$--$1.031$ (crest) and $0.969$--$1.042$ (SAG) of
static in CarSim, and our model reproduces these extremes within $0.002$.
Full-trace total-pitch RMSE is $0.058^\circ$ (crest) and $0.051^\circ$ (SAG).
Nearly all of it is a constant $0.052^\circ$ static-equilibrium offset that is
present before the road event; without it the RMSE is $0.012^\circ$ and
$0.016^\circ$.

\begin{figure}[t]
\centering
\includegraphics[width=\linewidth]{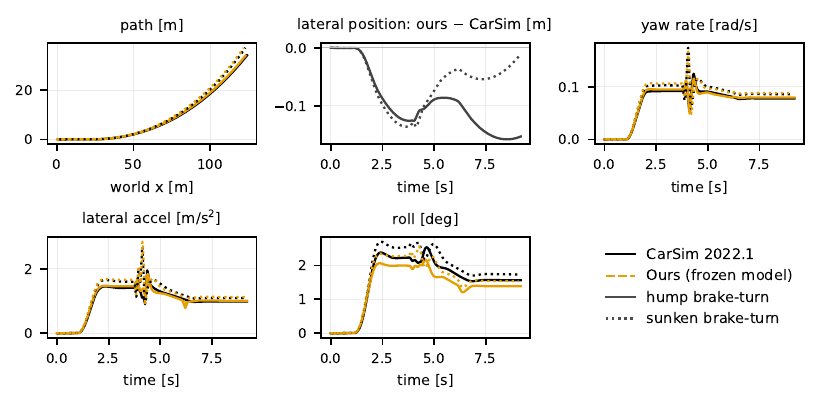}
\caption{Lateral response on two steered holdouts (solid: hump with braking and
turning; dotted: sunken road with braking and turning). Black is CarSim and orange
is our model. The second panel shows our model minus CarSim.}
\label{fig:carsim_lateral_steered}
\end{figure}

\paragraph{Lateral response.}
The crest and SAG runs have zero steering, so lateral evidence comes from the
steered cases (Figure~\ref{fig:carsim_lateral_steered}). The front lateral tire-force
scale ($0.85$) was fitted on two development brake-turn cases (hump and sunken
road); speed, steering, and wheel torque are replayed identically in every case.
Table~\ref{tab:lateral_recal} gives the holdout accuracy: lateral position RMSE is
$0.03$--$0.11\,\mathrm{m}$ and heading RMSE at most $6.3\,\mathrm{mrad}$, and all
gates and all $36$ convergence checks pass. The three steered holdouts were evaluated only after the final model
parameters had been frozen (role E). The development cases outside the fit, the flat-road
steering step and the low-friction brake-turn, have lateral position RMSE $0.16$ and $0.09\,\mathrm{m}$.

Residual biases remain. At steady state yaw rate and lateral acceleration are
$3$--$4\%$ above CarSim, and the small end drift ($0.02$--$0.15\,\mathrm{m}$ after
$9.2\,\mathrm{s}$) comes from yaw-rate errors of alternating sign
(over-response in the steady phases, under-response during the braking
transient). On the sinusoidal-steer holdout the yaw-rate NRMSE is $0.032$, well
inside its limit of $0.20$ but higher than on the brake-turns. Roll angle is $11\%$
below CarSim and roll per unit lateral acceleration is $14\%$ lower, which places
this bias in the roll submodel rather than the tire.

\begin{figure}[t]
\centering
\includegraphics[width=\linewidth]{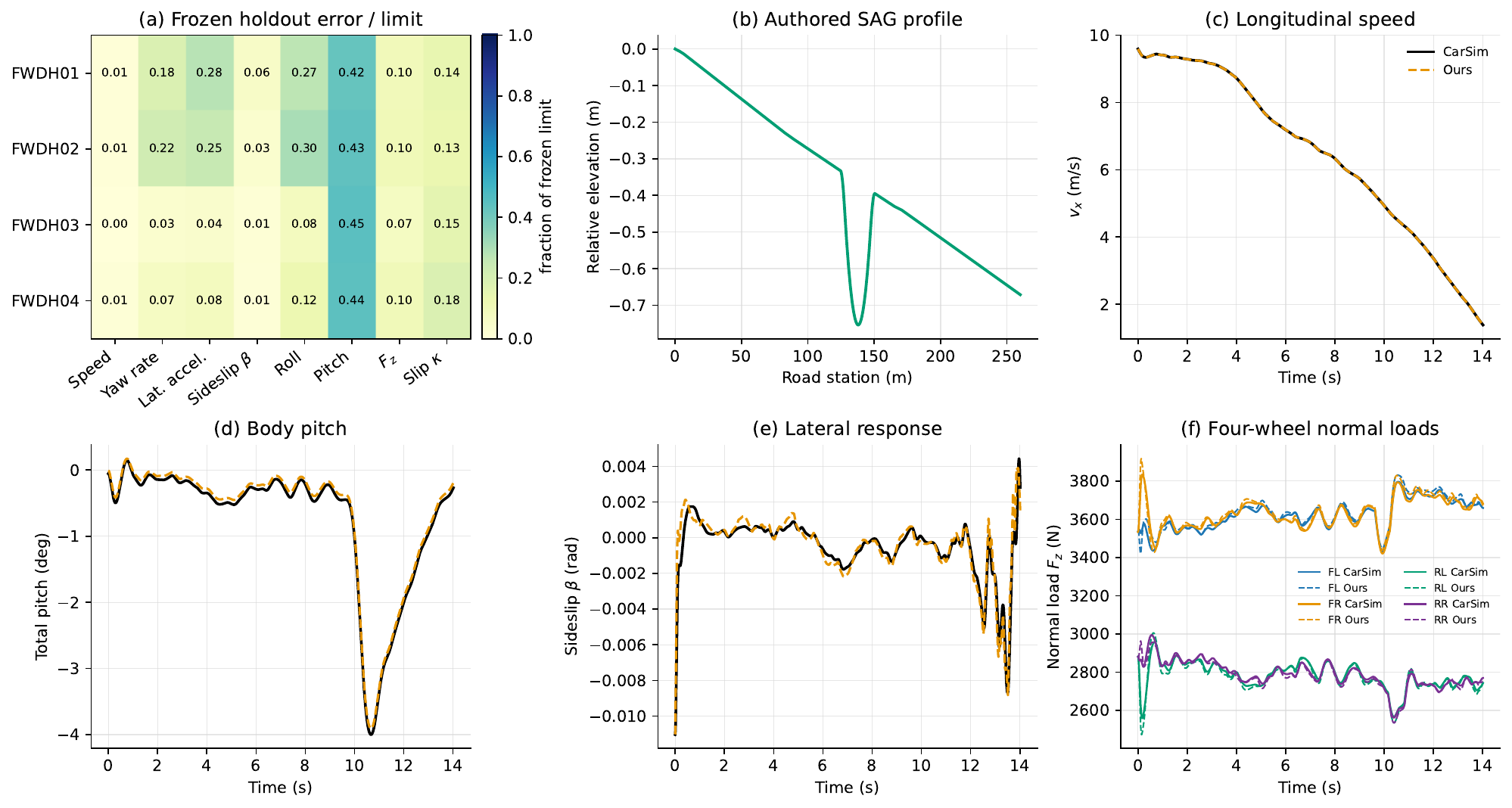}
\caption{Scene-supported SAG holdout. (a) Gate margin (error over limit) for
all four cases and eight metrics. (b)--(f) Case 2: authored SAG profile and
raw-time agreement in speed, pitch, sideslip, and four-wheel normal loads.}
\label{fig:carsim_forward_holdout}
\end{figure}

\paragraph{Scene-supported SAG holdout.}
Four cases were selected before their traces were generated. They span SAG grade,
friction, road length, initial state, braking, and steering. All pass every
integrity and dynamics gate, including the lateral-state checks, all $192$
initialization gates, and all $72$ timestep-convergence gates
(Table~\ref{tab:carsim_forward_holdout}, Figure~\ref{fig:carsim_forward_holdout}).
The suite covers these four combinations under the rules $v_x\geq0.5\,\mathrm{m/s}$,
no reverse motion, positive wheel contact, and bounded combined tire utilization.

\begin{table}[t]
\centering
\caption{RMSE over the full horizon on the four scene-supported SAG cases.}
\label{tab:carsim_forward_holdout}
\small
\begin{tabular}{lrrrrrrr}
\hline
Case & $G$ (\%) & $\mu$ & $L$ (m) & $v$ (m/s) & $\theta$ ($^\circ$) &
$\beta$ (rad) & $F_z$ (N) \\
\hline
1 & 4 & 0.55 & 59.74 & 0.00077 & 0.0636 & 0.00083 & 14.59 \\
2 & 6 & 0.40 & 20.00 & 0.00080 & 0.0652 & 0.00050 & 15.32 \\
3 & 2 & 0.70 & 59.74 & 0.00019 & 0.0669 & 0.00020 & 11.11 \\
4 & 5 & 0.85 & 74.68 & 0.00118 & 0.0667 & 0.00016 & 15.06 \\
\hline
Mean & -- & -- & -- & 0.00073 & 0.0656 & 0.00042 & 14.02 \\
\hline
\end{tabular}
\end{table}

\begin{figure}[t]
\centering
\includegraphics[width=\linewidth]{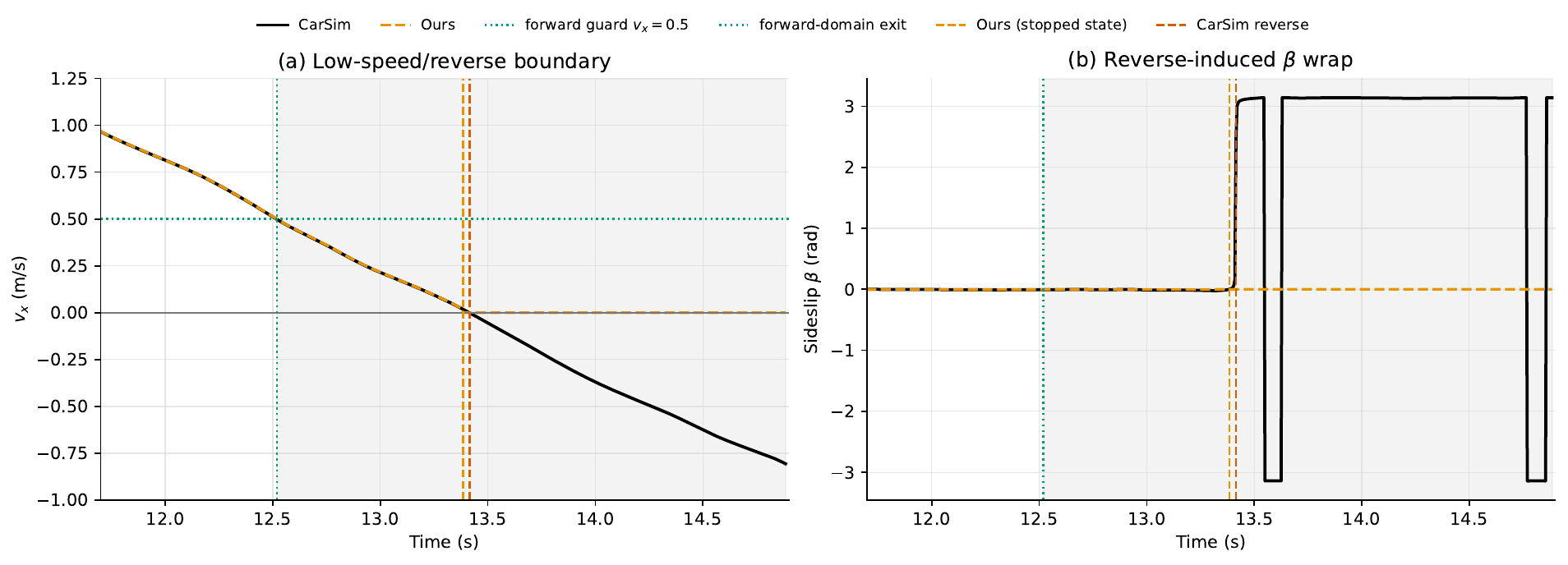}
\caption{The failed stress case. Traces agree until both simulators leave the
forward-motion domain; our model then holds the vehicle at rest while CarSim
reverses, and the resulting reverse-motion sideslip wrap explains the
full-horizon metric failure.}
\label{fig:carsim_sagq05_failure}
\end{figure}

\paragraph{SAG stress and the reverse-motion boundary.}
Six further SAG cases cover grade, curvature, friction, initial-state, braking,
and steering boundaries. Five pass every gate. The sixth fails only after the
trajectory leaves the forward-motion domain: both simulators fall below the
$0.5\,\mathrm{m/s}$ guard at about $12.52\,\mathrm{s}$; our model then holds the
vehicle at rest, whereas CarSim crosses zero speed and reverses to about
$-0.81\,\mathrm{m/s}$. The full-horizon speed RMSE ($0.157\,\mathrm{m/s}$, limit
$0.12$) and sideslip RMSE ($1.003\,\mathrm{rad}$, limit $0.015$) then exceed
their limits. Initialization and convergence gates all pass, so this is a
model-boundary mismatch, not a numerical one, and we report it as a failure
(Figure~\ref{fig:carsim_sagq05_failure}).

\paragraph{Lateral response of the holdouts.} Table~\ref{tab:lateral_recal} gives the accuracy on the three holdouts of
Section~\ref{sec:carsim_validation}, and Figure~\ref{fig:carsim_lateral_steered} shows the two brake-turn cases.

\begin{table}[t]
\centering
\caption{Lateral accuracy on the three holdouts (raw $5\,$ms traces, full
horizon). The last row gives the pass gate.}
\label{tab:lateral_recal}
\scriptsize
\begin{tabular}{l l l l l}
\hline
Case & $y$ RMSE [m] & Heading RMSE [mrad] & Yaw-rate NRMSE & Roll NRMSE \\
\hline
Sunken, brake-turn & $0.075$ & $6.3$ & $0.017$ & $0.095$ \\
Hump, brake-turn & $0.110$ & $4.3$ & $0.022$ & $0.093$ \\
SAG, sinusoidal steer & $0.035$ & $1.2$ & $0.032$ & $0.096$ \\
\hline
Gate & $1.5$ & $50$ & $0.20$ & $0.25$ \\
\hline
\end{tabular}
\end{table}

\section{Additional ZOD results}
\label{app:zod}

\begin{figure}[t]
\centering
\includegraphics[width=\linewidth]{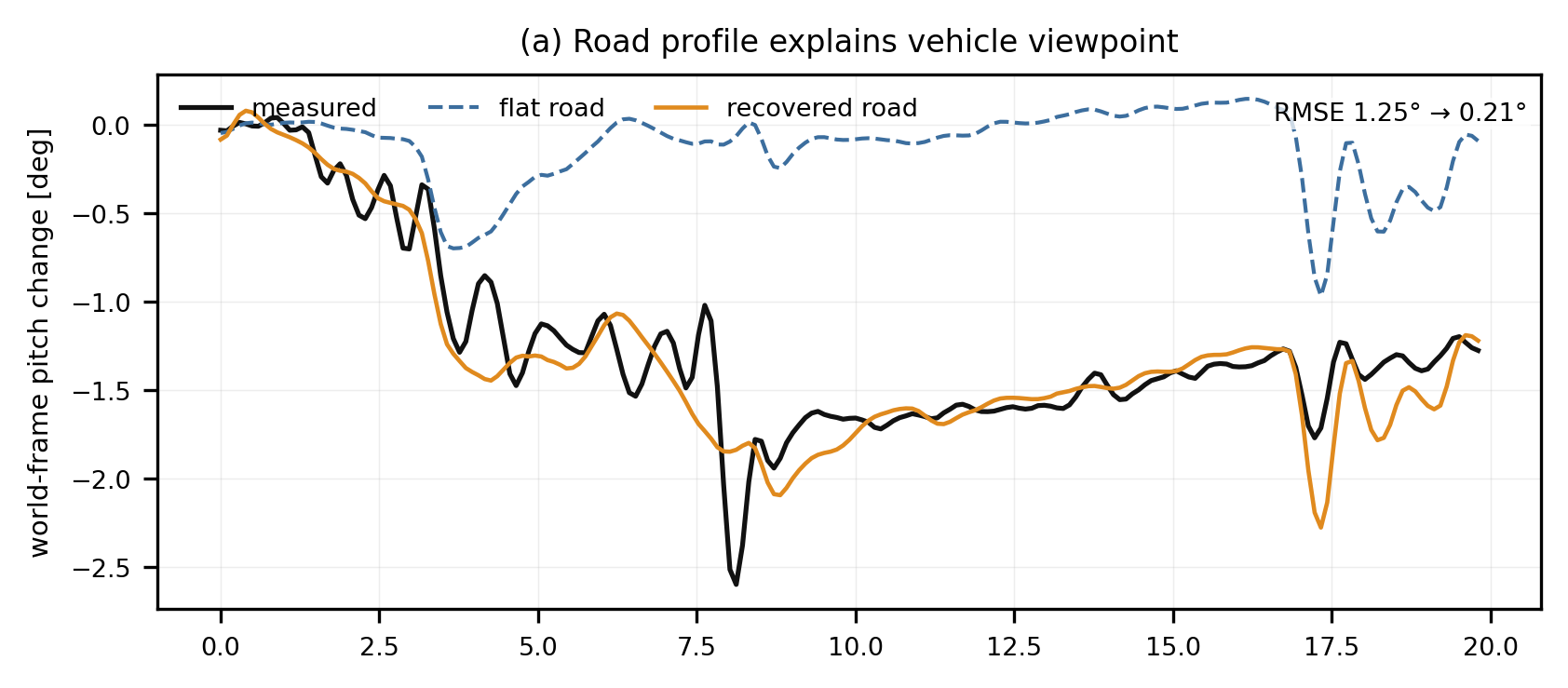}
\caption{Measured vehicle world-frame pitch and predictions from a flat-road
assumption and the recovered road profile on a held-out ZOD sequence. The LiDAR
geometry check is in Figure~\ref{fig:zod_lidar}.}
\label{fig:zod_profile}
\end{figure}

\paragraph{The test sequence shown in Figure~\ref{fig:zod_profile}.} Over its first ten seconds the grade falls from $2.9\%$ to zero and the speed from $39$ to $26\,\mathrm{km/h}$ over the whole sequence, while the measured pitch changes by about $1.7^\circ$, with a $2.6^\circ$ excursion at a braking event. Speed and grade are inputs to the predictions, not outputs, so they are not compared.

\paragraph{Protocol.} Measured speed and longitudinal and lateral acceleration condition the rollout of the half-car inherited from muSync-GS with its CarSim-identified parameters (not the coupled vehicle model); the model predicts body pitch and heave, which are compared with the OXTS pitch and height signals. Three predictions are compared: a flat road, a \emph{kinematic} camera that follows the recovered road's height and pitch with no suspension, and the \emph{full} half-car on the recovered road. No parameter is fitted on ZOD; only a static per-event sensor offset is removed, and no time alignment is applied. Events are selected from road geometry and motion support alone, before any response is inspected: $29$ drives from $13$ independent recordings give $191$ candidate events, of which $96$ recording-balanced events (at most two per recording and road type: uphill, downhill, crest, SAG, curve with grade, flat) are evaluated. Errors are averaged over recordings rather than events, so that a recording with many events does not dominate. Recordings are pre-assigned to train, validation, and test ($7/3/3$); because nothing is fitted, we report the pre-declared test split and all $13$ recordings together. With a front-camera focal length of about $1.87\times10^{3}\,\mathrm{px}$, one degree of pitch shifts the image by about $33\,\mathrm{px}$. The world-frame pitch of the vehicle sets the viewing direction of the camera, so it is the quantity on which the road input matters most for rendering. The $95\%$ cluster-bootstrap interval of the full model's world-frame pitch RMSE on the test recordings is $0.13$--$0.47^\circ$, against $0.31$--$2.30^\circ$ for the flat baseline.

\begin{figure}[t]
\centering
\includegraphics[width=\linewidth]{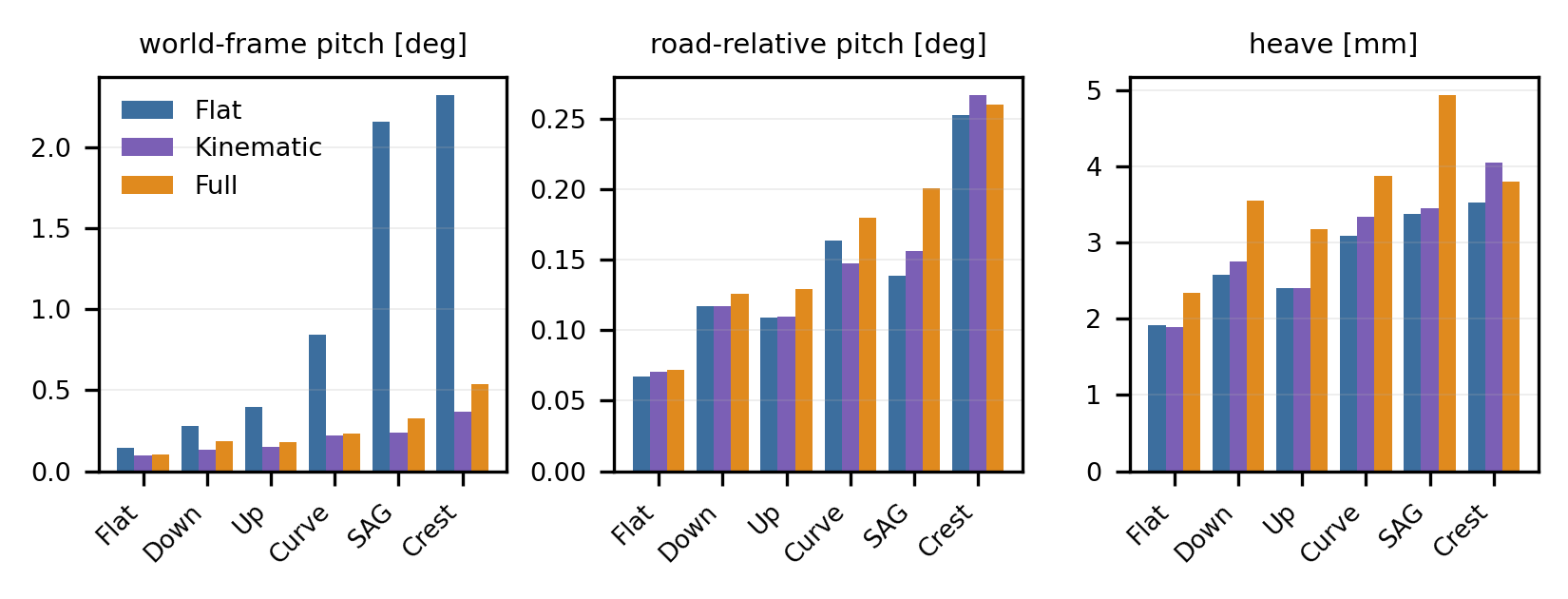}
\caption{ZOD error (RMSE) by event type for world-frame pitch, road-relative pitch, and
heave, averaged over the recordings that contain each type ($10$, $10$, $12$, $12$, $5$,
and $3$ recordings for flat, downhill, uphill, curve with grade, SAG, and crest). Types
come from road geometry and were fixed before any response was inspected.}
\label{fig:zod_types}
\end{figure}

\paragraph{All three quantities by event type.} Figure~\ref{fig:zod_types} shows the world-frame pitch together with
road-relative pitch and heave, and Table~\ref{tab:zod} gives the RMSE of the three predictions on the pre-declared test
split and on all $13$ recordings. In world-frame pitch the road input removes most of the error where the road has
vertical curvature. In road-relative pitch and heave the three predictions are close in every event type, the flat
baseline is never worse than the full model, and all methods err more on the harder event types.

\begin{table}[t]
\centering
\caption{ZOD prediction error (RMSE), averaged over recordings. Test is the
pre-declared split of $3$ recordings; All pools all $13$.}
\label{tab:zod}
\small
\begin{tabular}{l rrr rrr}
\hline
 & \multicolumn{3}{c}{Test (3)} & \multicolumn{3}{c}{All (13)} \\
Quantity & Flat & Kin. & Full & Flat & Kin. & Full \\
\hline
World-frame pitch ($^\circ$) & 1.027 & 0.236 & 0.258 & 0.594 & 0.160 & 0.196 \\
Road-relative pitch ($^\circ$) & 0.140 & 0.145 & 0.168 & 0.120 & 0.119 & 0.136 \\
Heave (mm) & 2.35 & 2.49 & 3.11 & 2.60 & 2.70 & 3.42 \\
\hline
\end{tabular}
\end{table}

\begin{figure}[t]
\centering
\includegraphics[width=\linewidth]{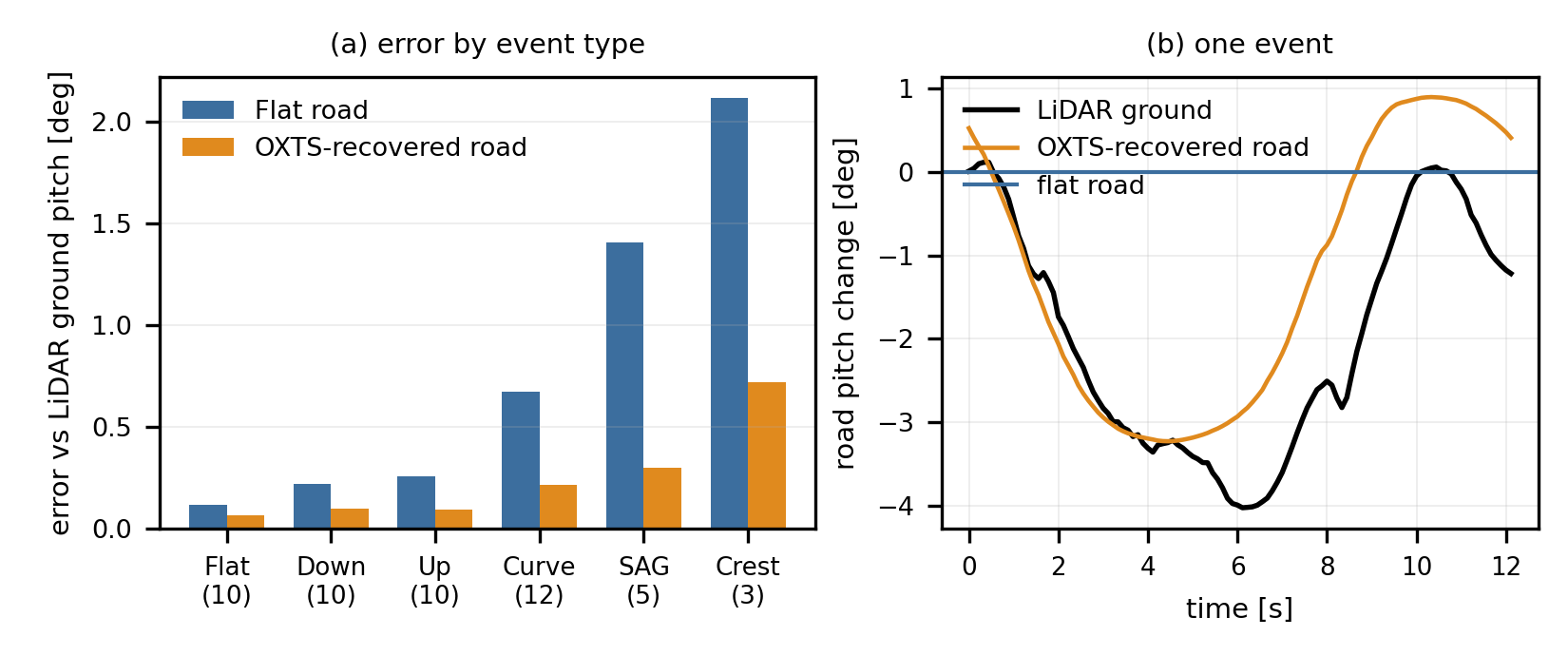}
\caption{The OXTS-recovered road against an independent LiDAR ground plane. (a) Error
of the flat-road assumption and of the OXTS-recovered road relative to the LiDAR
ground pitch, by event type (recordings per type in brackets); LiDAR covers $79$ of
the $96$ events. (b) The test event with the highest geometry score that has LiDAR.}
\label{fig:zod_lidar}
\end{figure}

\paragraph{Independent check with LiDAR.}
The OXTS profile and the measured pitch come from one inertial unit. To test the
recovered road against an independent source, we fitted a ground plane to the LiDAR
points on the lane $5$--$15\,\mathrm{m}$ ahead of the vehicle in every scan inside an
event (about nine scans per second). This uses neither altitude nor GNSS. Adding the
measured world pitch of the vehicle gives the road pitch in the world frame, which we
compare with the OXTS-recovered road pitch at the same point, with the same
one-second baseline as above; the patch and thresholds were fixed before any result
was computed. LiDAR is available for $79$ of the $96$ events ($13$ recordings); the
rest had too few valid ground fits. Against this reference the flat-road assumption
has an error of $0.45^\circ$ (average over recordings) and the OXTS-recovered road
$0.14^\circ$, a reduction of $68\%$, in line with the $67\%$ obtained against the
measured pitch. The pattern again follows vertical curvature
(Figure~\ref{fig:zod_lidar}a): crest $2.11^\circ$ against $0.72^\circ$, SAG
$1.40^\circ$ against $0.30^\circ$, and curves with grade $0.67^\circ$ against
$0.21^\circ$. LiDAR alone, without any pose, shows the same thing: on crest and SAG
events the ground tilts relative to the vehicle by only $0.84^\circ$ and $0.32^\circ$
(RMS, event means) while the vehicle's own pitch changes by $2.30^\circ$ and
$1.49^\circ$, so the ground rotates with the vehicle.
The LiDAR ground fit has a noise of about $0.14^\circ$, which bounds the pitch
change it resolves; the crest error is $0.72^\circ$, and the event in
Figure~\ref{fig:zod_lidar}b, chosen by a pre-declared rule (highest geometry score
among the test events with LiDAR), agrees for the first five seconds and then
differs by up to about $2^\circ$.

\section{Additional geometry results}
\label{app:geometry}

\begin{figure}[t]
\centering
\includegraphics[width=\linewidth]{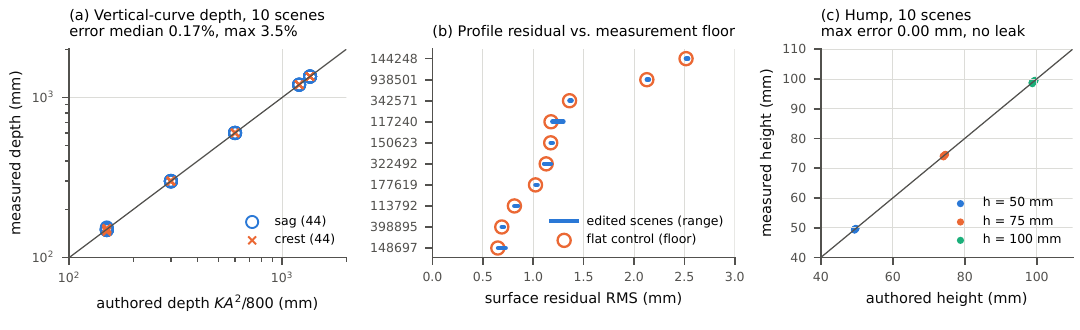}
\caption{Authored against recovered road geometry in the deformed Gaussian scenes of
ten Waymo scenes. (a) Apex depth of $88$ vertical curves. (b) Residual RMS of the
recovered profile for each scene (range over its edits) against the same scene with a
flat edit, which sets the floor of the measurement. (c) Height of $30$ humps.}
\label{fig:geometry_fidelity}
\end{figure}

Figure~\ref{fig:geometry_fidelity} shows the authored against the recovered geometry of the deformed scenes of
Section~\ref{sec:geometry_fidelity}; the setup and results follow.

The dynamics and the camera read the edited profile $H_{0}(s)+\Delta H_{\eta}(s)$ analytically, so
they receive it by construction. The deformed Gaussian scene does not: it is the
output of the attachment and deformation rules of Eq.~\eqref{eq:gaussian_deform}.
We therefore ask whether a vertical curve authored with apex depth $X$ is $X$ deep in
the edited scene, measured in the scene itself and without rendering.

\paragraph{Setup.} We use all ten reconstructed Waymo scenes, with routes of
$70$--$263\,\mathrm{m}$, including native grades of $+16\%$, $-11\%$, and $-9\%$ and
a curved route. Each scene receives a single symmetric crest and a single symmetric SAG
centerd on the route, with algebraic grade change $A\in\{4,8,12\}\%$ and
$K=L/A\in\{7.5,15\}\,\mathrm{m}$ per percent, and raised-cosine humps of
$50$, $75$, and $100\,\mathrm{mm}$ ($3.66\,\mathrm{m}$ long). Curves that do not fit on the
route with $10\,\mathrm{m}$ to spare at each end are recorded as infeasible and not
altered, which leaves $88$ vertical curves (apex depths $K A^{2}/800$ of
$0.15$--$2.70\,\mathrm{m}$) and $30$ humps. Edits are added to the native profile, and
the measured quantity is edited minus original, so a native grade cancels. The scene is
deformed with the default pipeline settings and no per-scene tuning. Road Gaussians
within $4\,\mathrm{m}$ of the route centerline are binned in $2\,\mathrm{m}$ of station,
and each bin's median height gives the recovered profile. The same resampling applied
with a flat edit gives a per-scene measurement floor.

\paragraph{Results.} All $118$
edits were deformed and measured without failure or warning
(Figure~\ref{fig:geometry_fidelity}). Across the $88$ vertical
curves the apex depth error has a median of $0.17\%$ (90th percentile $0.86\%$,
maximum $3.5\%$), the vertical-curvature error, which drives the load term of the
dynamics, a median of $0.05\%$ (maximum $2.8\%$), and the grade error at most
$0.06$ percentage points. The residual surface error (median RMS $1.2\,\mathrm{mm}$,
maximum $2.5\,\mathrm{mm}$) is the measurement floor: on every scene it exceeds the flat
edit's residual by at most $0.10\,\mathrm{mm}$, and the largest depth and curvature
errors occur on the scene with the largest floor. Per Gaussian, every road Gaussian in
the lateral core of the edit is displaced by the authored amount to within
$0.06\,\mathrm{mm}$. The humps are applied to every road Gaussian in their footprint with
zero residual, and no Gaussian outside the footprint moves. Non-road Gaussians inside a
hump footprint occur on two scenes; all are tree canopy $2.3$--$7.4\,\mathrm{m}$ above
the road, which by design does not follow a hump, and none lies within
$0.3\,\mathrm{m}$ of the road surface.

That the rendered pixels follow the same profile is examined through the camera in
Section~\ref{sec:generators}, where the far-field shift is approximately linear in the
authored road pitch for the crest and sag camera paths.

\section{Additional multi-vehicle results}
\label{app:multivehicle}

\subsection{Twenty-episode bank}
\label{app:mv_bank}

The bank tests, on episodes never used elsewhere in this paper, whether the propagation seen on
109409 (Appendix~\ref{app:mv_109409}) generalizes, and whether letting surrounding vehicles respond
changes the ego's estimated interaction. Nothing is rendered. Every threshold, the selection seed,
and the placement rule were fixed in a configuration file before any edited run.

\paragraph{Population and selection.} The population is $3{,}791$ Waymo Open Motion training-split
episodes of $20\,\mathrm{s}$ that our factual pipeline had already audited and serialized. An
episode is eligible if (i) its serialized package exists ($789$ fail), (ii) it passes the package's
geometric audit, (iii) it has a factual ego--lead association in the package ($2{,}657$ fail),
(iv) the ego route is at least $150\,\mathrm{m}$, the $120\,\mathrm{m}$ edit plus $15\,\mathrm{m}$ at
each end ($2{,}294$ fail), (v) it has at least four moving vehicles including the ego, and (vi) it
is not used by any other experiment here ($65$ fail); criteria overlap, and $394$ episodes are
eligible. Their sorted ids are permuted once with a fixed seed and the first $20$ are taken.

\paragraph{Vehicles and validity.} Within an episode, a tracked vehicle is admitted if it is
observed for at least $3\,\mathrm{s}$, travels at least $2\,\mathrm{m}$, and comes within
$30\,\mathrm{m}$ of the ego route, up to $24$ per episode by ascending track id; $387$ are admitted.
A rollout is valid if control recovery (Section~\ref{sec:grounding}) completes; the $34$ that do not
all carry one reason, a recorded track that stops and restarts, which the model cannot relaunch
from rest, and are refused before any edit. This leaves $353$ valid vehicles (ego included).
Factual replay has median speed RMSE $0.018\,\mathrm{m/s}$ (90th percentile $0.028$). On a zero edit
every valid vehicle reproduces its baseline exactly. Pedestrians, cyclists, and parked vehicles are
not simulated.

\paragraph{Edit.} A bounded symmetric crest or sag with $30/60/30\,\mathrm{m}$ entry, main, and exit
curves returns to the native road at both ends and is centred on the ego route. Arms: $2\%$
($K=15\,\mathrm{m}$ per percent, apex offset $0.60\,\mathrm{m}$; primary) and $8\%$ ($K=3.75$,
$2.40\,\mathrm{m}$; stress), at $\mu=1$. Each vehicle queries the edit through its own unrolled
road; steering is zero and the plan-view path is replayed, so lateral motion cannot change.

\paragraph{Response.} A vehicle \emph{traverses} the edit if a front- or rear-axle road query of its
recorded track falls inside the edit support at any valid frame; this is decided before any
rollout. It \emph{responds} if its peak $|\Delta v|\ge0.05\,\mathrm{m/s}$ or its peak change of total
normal load is at least $0.005$ of static, each against its own flat-road rollout. Onset delay is
measured from its first frame inside the support on the $0.1\,\mathrm{s}$ output clock.

\paragraph{Pairing.} Each episode contributes at most one pair: the ego and the package's factual
lead. With $g$ their along-path gap and subscripts $\eta$ and $0$ for the edited and flat-road
rollouts,
\begin{equation}
\Delta g_{\rm all}=g(\text{ego}_\eta,\text{lead}_\eta)-g(\text{ego}_0,\text{lead}_0),\qquad
\Delta g_{\rm ego}=g(\text{ego}_\eta,\text{lead}_0)-g(\text{ego}_0,\text{lead}_0),
\label{eq:ego_only}
\end{equation}
so $\Delta g_{\rm ego}$ is the estimate obtained if only the ego responds and the lead keeps its
flat-road motion, which for a zero edit is its recording. The pre-declared statistic is
$D=\max_t|\Delta g_{\rm all}-\Delta g_{\rm ego}|$ over frames where both vehicles are valid, with
thresholds $0.5$ and $1\,\mathrm{m}$. One episode has no valid ego. Post hoc, an episode is excluded from every arm if
either vehicle's edited run is latched at rest by the model in any arm (the edit slows a vehicle the
recording kept moving until the model can no longer relaunch it, producing a large spurious change);
the flag uses the model's own thresholds, not the outcome size, and fires on one episode, in both sag
arms. This leaves $18$ pairs in every arm. The sign of the disagreement is reported only as a diagnostic, because which way a
gap moves depends on the assumed driver.

\paragraph{Statistics.} Episodes are the resampling unit ($2{,}000$ bootstrap resamples, fixed seed,
$95\%$ percentile intervals); vehicles within an episode are not independent.

\begin{figure}[t]
\centering
\includegraphics[width=\linewidth]{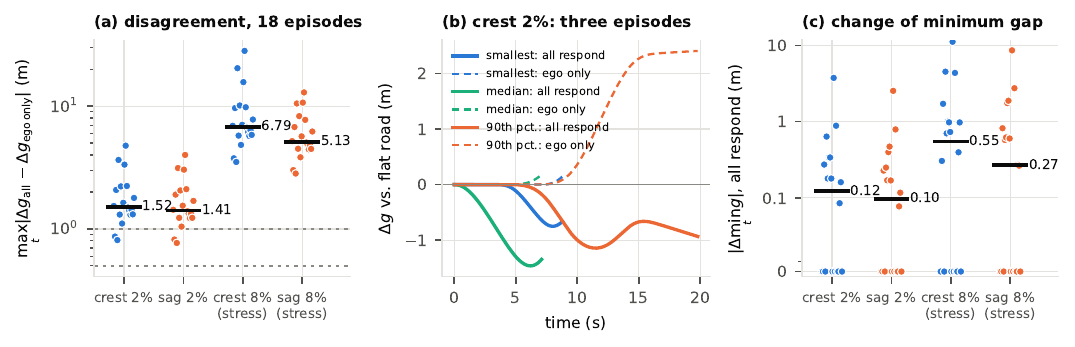}
\caption{All vehicles responding versus the ego alone, on the bank's ego--lead pairs. (a) $D$, one
point per episode; bars: medians; dotted: $0.5$ and $1\,\mathrm{m}$. (b) Gap change over time for three episodes chosen by a rule
fixed before plotting (smallest non-zero, median, and 90th-percentile $D$ at crest $2\%$); solid:
all respond, dashed: ego only. (c) Change of the minimum gap when all vehicles respond
(symmetric-log axis; the row at zero is episodes whose closest approach precedes the edit).}
\label{fig:bank_interaction}
\end{figure}

\paragraph{Results.} Table~\ref{tab:mv_bank} and Figure~\ref{fig:bank_interaction} give the result. At
$2\%$, $257$ of $266$ traversing vehicles respond and all $87$ non-traversers show exactly zero
change. The latter holds by construction, since the edit depends on station alone, and verifies
that the traversal test and the solver agree. Restricted to vehicles on
the ego's road, $97.5\%$ [$95.7$, $99.4$] of traversers respond and $0$ of $46$ non-traversers
change. Onset is one output frame at the median. The ego--lead disagreement $D$ exceeds
$0.5\,\mathrm{m}$ in every episode at both severities. The minimum gap changes far less, and not at
all in $8$ of $18$ episodes whose closest approach precedes the edit. The 109409 episode is re-run by the same code as a regression
test and reproduces its published numbers.

\begin{table}[t]
\centering
\caption{Twenty-episode bank. Response rows: $353$ valid vehicles in $20$ episodes. Pair rows:
$18$ ego--lead pairs, the same episodes in every arm. $D$: largest
disagreement between the all-vehicles and ego-only gap changes (Eq.~\ref{eq:ego_only}). Brackets:
$95\%$ episode-bootstrap intervals.}
\label{tab:mv_bank}
\scriptsize
\setlength{\tabcolsep}{3pt}
\begin{tabular}{l cccc}
\toprule
 & crest $2\%$ & sag $2\%$ & crest $8\%$ & sag $8\%$ \\
\midrule
Traversers responding & $257/266$ & $257/266$ & $261/266$ & $261/266$ \\
Non-traversers changed & $0/87$ & $0/87$ & $0/87$ & $0/87$ \\
Median peak $|\Delta v|$ of traversers (m/s) & $0.35$ & $0.34$ & $1.44$ & $1.35$ \\
Median onset delay (s) & $0.1$ & $0.1$ & $0.1$ & $0.1$ \\
\midrule
Median $D$ (m) & $1.52$ [$1.31$, $2.08$] & $1.41$ [$1.23$, $1.90$] & $6.79$ [$5.84$, $9.65$] & $5.13$ [$4.49$, $7.25$] \\
90th percentile of $D$ (m) & $3.43$ & $3.07$ & $17.11$ & $10.57$ \\
Pairs with $D>0.5$ / $>1\,$m & $100\%$ / $89\%$ & $100\%$ / $89\%$ & $100\%$ / $100\%$ & $100\%$ / $100\%$ \\
Median $|\Delta\min_t g|$, all respond (m) & $0.12$ & $0.10$ & $0.55$ & $0.27$ \\
Pairs with minimum gap unchanged & $8$ & $8$ & $7$ & $8$ \\
\bottomrule
\end{tabular}
\end{table}

\subsection{The reconstructed episode 109409}
\label{app:mv_109409}

This is the rendered episode of Figure~\ref{fig:full_chain}. Every tracked moving vehicle is
rolled out on the edited road (Section~\ref{sec:traffic_objects}), and we ask whether each one
responds to the edit, when, and only if it drives over it. Unlike the bank of
Appendix~\ref{app:mv_bank}, it uses the $8\%$ stress edit, placed $35\,\mathrm{m}$ into the ego
route rather than centred on it.

\paragraph{Setup.} The scene has $17$ moving tracked vehicles besides the ego. Each is
grounded independently (Section~\ref{sec:grounding}): its wheel torque is identified
against its own recorded trajectory, which reproduces the recorded speed to
$0.008$--$0.034\,\mathrm{m/s}$ RMSE for all $18$ vehicles, and is then frozen. Each
vehicle sees the road through its own recorded path distance, so crossing and
oncoming vehicles are not forced onto the ego's station. The edit is a bounded
symmetric crest or SAG of $\pm8\%$ starting $35\,\mathrm{m}$ into the ego route
($K=3.75\,\mathrm{m}$ per percent, below the internal plausibility threshold used here
and therefore stress geometry), at $\mu=1.0$. The response of a vehicle is its edited rollout minus its
flat-road rollout, so a zero edit is an exact identity. The recorded plan-view path is
replayed, so this experiment exercises the longitudinal and vertical coupling, not the
lateral one; every vehicle uses the same chassis parameters. The five tracked vehicles that never move are replayed at their recorded poses, and the scene contains no tracked pedestrians or cyclists (Table~\ref{tab:object_scope}).

\begin{figure}[t]
\centering
\includegraphics[width=\linewidth]{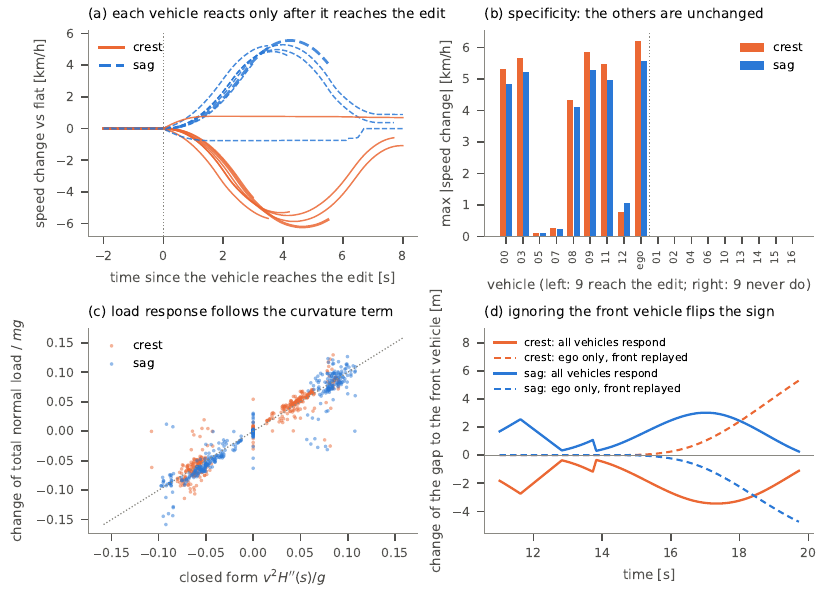}
\caption{Every vehicle of a real Waymo drive on the edited road. (a) Speed change of
each vehicle that reaches the edit, aligned at the time it does. (b) Largest speed
change per vehicle; the nine on the right never reach the edit. (c) Change of total
normal load against the closed-form curvature term $v^{2}H''(s)/g$, over all frames
inside the edit. (d) Change of the gap from the ego to the vehicle ahead when all
vehicles respond (solid) and when the vehicle ahead is replayed as recorded (dashed).}
\label{fig:multivehicle}
\end{figure}

\paragraph{Results.} Figure~\ref{fig:multivehicle} gives the result. Nine of the $18$
vehicles reach the edited stretch and nine never do; for the latter the speed and path
change is exactly zero. No vehicle departs from its flat-road rollout before it reaches
the edit or the counterfactual onset, and each changes speed within one $0.1\,\mathrm{s}$
frame of reaching it. The affected vehicles change speed by up to $6.2$ (crest) and
$5.6\,\mathrm{km/h}$ (SAG) and path distance by up to $7.3$ and $5.8\,\mathrm{m}$, and
their total normal load spans $0.90$--$1.13$ and $0.85$--$1.13$ of static. Over the
$451$ frames inside the edit the load change follows the closed form $v^{2}H''(s)/g$
with correlation $0.95$ (crest) and $0.96$ (SAG) and slope $0.99$ and $1.00$, with a
mean absolute difference of $0.009$ and $0.011$ of the static weight; the remaining
difference sits at the curvature steps, where the model spreads over the wheelbase what
the closed form applies at the center of mass.

Letting the other vehicles respond changes the interaction the ego sees. The vehicle ahead reaches the edit first, so with all moving vehicles responding the crest shortens the along-path gap to it by up to
$3.4\,\mathrm{m}$ and the SAG lengthens it by up to $3.0\,\mathrm{m}$, whereas a rollout that
moves only the ego and replays the vehicle ahead as recorded gives $+5.3$ and
$-4.7\,\mathrm{m}$, the opposite sign in every frame where the change exceeds
$0.5\,\mathrm{m}$. The minimum gap itself changes by less than $0.5\,\mathrm{m}$
($30.1\,\mathrm{m}$ flat).

\subsection{A prescribed-input case}
\begin{figure}[t]
\centering
\includegraphics[width=0.9\linewidth]{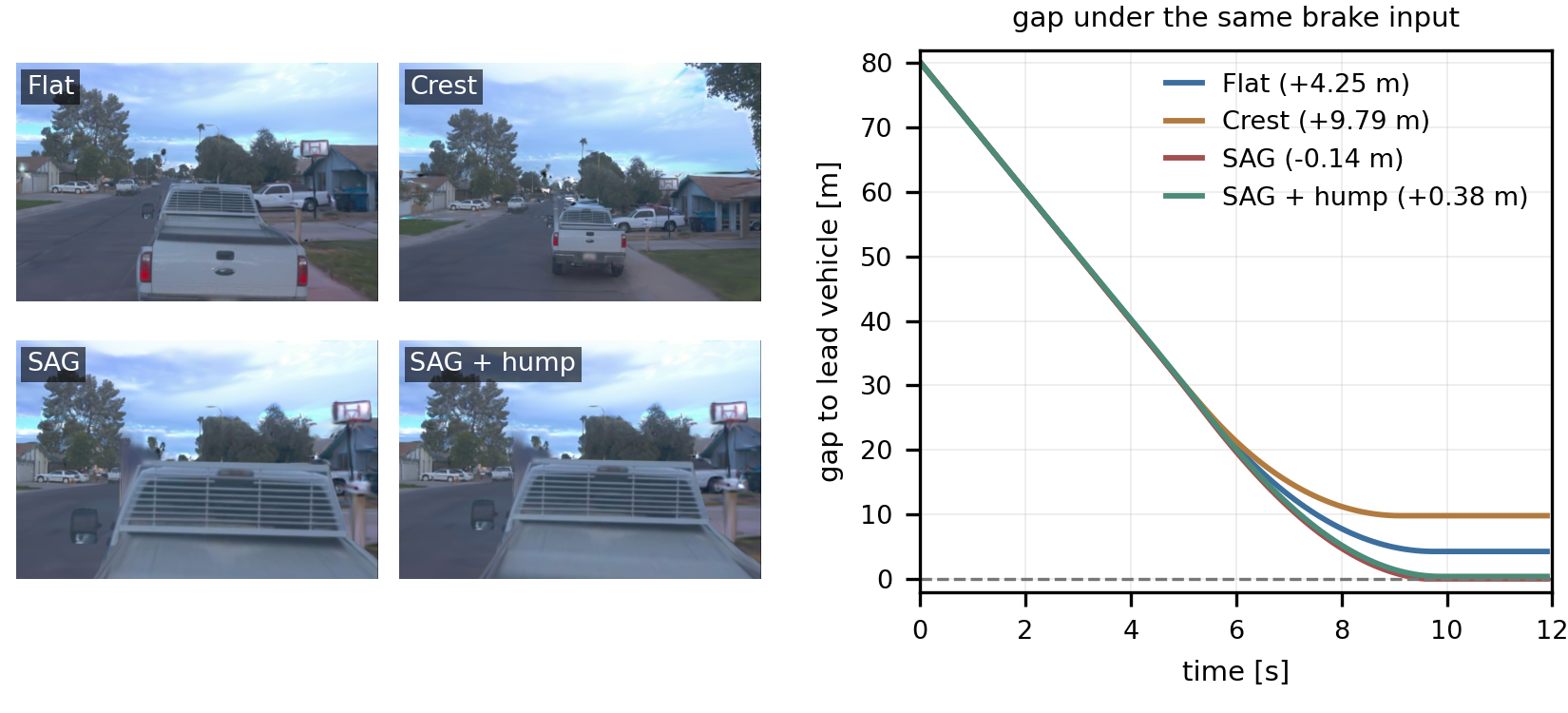}
\caption{Prescribed-input case on Scene 22 (ego only, static lead vehicle). Left:
the same rendered instant ($t=9.5\,\mathrm{s}$) under the four road arms, from the
dynamic camera. Right: gap to the lead vehicle under the same prescribed input
($36\,\mathrm{km/h}$, $1000\,\mathrm{N\,m}$ brake from $5.34\,\mathrm{s}$, onset chosen
to expose the difference); the legend gives each arm's minimum ego--lead gap. Gaps are computed on the
metric rollout poses.}
\label{fig:synchronized_case}
\end{figure}

Figure~\ref{fig:synchronized_case} uses prescribed rather than inferred controls (Table~\ref{tab:setup_extra}): on Scene 22 the ego starts at $36\,\mathrm{km/h}$ and brakes with the same command on four roads toward a static lead vehicle. The brake onset was chosen to expose the braking mechanism.

\section{Generic-generator comparison: protocol and additional results}
\label{app:generators}

\paragraph{Protocol and measurement.}
\paragraph{Setup.} The text-only arms start from the Scene~22 clip used in the main synchronized example: 81
frames at $10\,\mathrm{fps}$ and $480\times320$, rendered from the flat reconstruction.
The edit is given in road coordinates. A crest or sag runs from $s=12$ to
$58\,\mathrm{m}$ with entry and exit grades of $\pm8\%$ and its apex or low point at
$s=35\,\mathrm{m}$; a sustained uphill or downhill covers the same interval; and the
interval is tied to frames 13, 41, and 78 of the clip. Three open models receive the
flat clip and this description and nothing else: VACE
(Wan2.1-VACE-14B~\citep{jiang2025vace,wan2025}) with a road mask derived from the input,
Cosmos-Transfer2.5-2B~\citep{nvidia2025cosmosphysicalai} with a blur control computed
from the input clip, and LTX-Video~0.9.8 13B~\citep{hacohen2024ltxvideo} in
video-to-video mode. In this arm none receives target geometry, depth, or a reference frame. Each
cell is one run at seed $2025$ with settings fixed before the run. The frames shown are
sampled at fixed road stations. The sag and crest are evaluated on three scenes with
the same settings (Waymo 150623, 177619, and 398895). Our arm authors the same profile in the road field,
deforms the scene, and re-renders it from a camera that follows the road height and
pitch on the original station schedule (the kinematic camera; the suspension response
of the rollout is not applied to these frames). The vehicle response comes from the
frozen model with the recorded speed profile as driver intent and $\mu=0.30$.
The frames and the response panel therefore come from two separate settings rather than one
rollout of the pipeline of Section~\ref{sec:multivehicle}: the frames test whether the
prescribed geometry appears in the image, and the response uses speed tracking
(Table~\ref{tab:setup}), which regenerates drive and brake torque on each road and so
asks what holding the recorded speed costs rather than how fixed actions play out.

\paragraph{Conditioned arms.} On Waymo 150623 we also give the generators structure about the
target road. \emph{Projected lines} are the route edges ($\pm1.8\,\mathrm{m}$) and a dashed
center line, projected through the camera of the edited scene; they come from the route, not
from a measured map. \emph{Target depth} is the depth video of the edited scene. For VACE it is
composited into the source inside the road mask. For Cosmos it is the depth control, either
inside the road mask with the source latents guided outside it (\emph{road only}), or over the
whole frame without guidance (\emph{full frame}) at control weights $0.3$, $0.6$, and $1.0$.
All of this structure is rendered from the TerrainForge-edited scene, so these arms evaluate
whether a generator can realize a supplied geometric scaffold. LTX has no geometry input, and VACE with full-frame depth was not run.

\paragraph{What is measured.} Generated RGB has no metric scale, so camera-trajectory
consistency is evaluated from recovered camera motion. A camera that follows a grade $g$ pitches by $\arctan g$, and the horizon moves in the
image accordingly. We measure the median vertical optical flow of textured pixels
above the flat-road horizon, against the flat render of the same frame for our arm and
against the input frame for the generators. Similarity to the input (SSIM) is reported as a
source-preservation diagnostic and is separate from the camera-trajectory criterion.

The far-field shift is an image-space quantity, so we add a second measurement on the recovered
camera trajectory. We recover the
camera path of each video with VGGT~\citep{wang2025vggt}, scale it to the authored station
span, and compare the camera height with the authored profile $H(s)$ by RMSE, correlation, and
the ratio of peak-to-peak heights. A recovered trajectory is \emph{consistent with the authored sag} if the RMSE is at
most $0.28\,\mathrm{m}$, the correlation at least $0.70$, and the ratio between $0.6$ and
$1.4$. The rule was fixed before any generator video was measured, from ten controls: our
renders of the sag on three scenes, flat renders on three scenes, and our renders of the crest,
uphill, downhill, and hump scored against the sag (Appendix~\ref{app:generators}). Pavement estimators did not separate the controls and are not used.

\paragraph{Results: text only.} Table~\ref{tab:generators} and Figure~\ref{fig:generators}, both in Appendix~\ref{app:generators}, give the
outcome on Scene~22. No generator produces the requested sag. VACE returns the input almost
unchanged: its lane region differs from the input by $2.9/255$ on average, the same as
adjacent input frames differ from each other ($2.8/255$), and its far-field shift is
below $0.1\,\mathrm{px}$. Cosmos and LTX regenerate the street, changing buildings,
vehicles, and vegetation (SSIM $0.69$ and $0.73$), and in none of the three does the
horizon move. The crest, uphill, and downhill requests give the same picture
(Table~\ref{tab:generators}): VACE returns the input (SSIM $0.94$ in every case), and
Cosmos and LTX regenerate the street ($0.69$--$0.70$ and $0.73$). The outputs of one
model for the four different edits are also nearly the same video: SSIM between edits is
$0.97$ for LTX and $0.86$--$0.95$ for Cosmos ($0.97$--$0.98$ for VACE, which copies the
input), against $0.73$ and $0.69$--$0.70$ to the input. How a generator departs from the
input is therefore set by the model and the seed, not by the requested edit. In our
arm the far field moves by $-47$ and
$+34\,\mathrm{px}$ at the steepest downhill and uphill positions of the sag
($s\approx20$ and $52\,\mathrm{m}$) and by $+45$ and $-34\,\mathrm{px}$ for the crest.
For the sag and the crest the shift is linear in the authored road pitch,
$8.8\,\mathrm{px}$ per degree for both (correlation $0.97$), which implies a focal length of about $500\,\mathrm{px}$
and a $51^\circ$ horizontal field of view at this resolution. The rendered camera
pitch equals $\arctan H'(s)$ to $0.00^\circ$, so this is a check that the render
follows the authored profile, not a second quantity. At the low point and apex the grade
is zero, and the remaining $\pm12\,\mathrm{px}$ is consistent with the camera sitting
$1.28\,\mathrm{m}$ lower or higher: parallax on trees and houses $20$--$60\,\mathrm{m}$
away is roughly $10$--$30\,\mathrm{px}$ at this focal length
(Appendix~\ref{app:generators}). For the sustained grades the far field moves by only
$+3$ and $-2\,\mathrm{px}$ at $s=20\,\mathrm{m}$, where the grade has just reached
$\pm7.8\%$, and by $+26$ and $-22\,\mathrm{px}$ at $36\,\mathrm{m}$. This is
consistent with the streetscape being sheared along with the road, which largely cancels
the camera pitch in the image: a sustained grade is much less visible than a crest or sag
of the same grade, and it shows up in the vehicle response instead, with the speed up to
$6.7\,\mathrm{km/h}$ below the flat road uphill and up to $3.7\,\mathrm{km/h}$ above
it downhill under the same driver intent (Appendix~\ref{app:generators}).

The three evaluated scenes give the same result for the sag and the crest
(Table~\ref{tab:generators_scenes}, Appendix~\ref{app:generators}). Measured against the input, the far field of every
generator output moves by at most $1.9\,\mathrm{px}$ in any frame, whereas ours moves by
$29$--$47\,\mathrm{px}$ at the steepest stations, with the sign set by the authored
grade. On these scenes VACE also changes the lane region more than adjacent input frames
differ, so the \emph{copy} label of Table~\ref{tab:generators} does not transfer and we
report only SSIM and the far-field shift.

\paragraph{Results: conditioned generators.} Figure~\ref{fig:generators_main} puts the arms side
by side. No text-only video is consistent with the target camera trajectory, on any of the three scenes ($0/9$), and
neither does any video given projected lines ($0/4$) or depth on the road only ($0/2$). Their
recovered camera height is flat (RMSE $0.45$--$0.53\,\mathrm{m}$, peak-to-peak ratio
$0.11$--$0.28$, against $0.44$--$0.52\,\mathrm{m}$ for the flat renders), and on 150623 their
far-field shift is at most $0.2\,\mathrm{px}$. Lines are drawn as texture and road-only depth is
ignored or painted as stripes. Only depth over the whole frame moves the video, and only at
some settings. At weight $0.6$ the recovered profile lies on the authored one (RMSE
$0.10\,\mathrm{m}$, correlation $0.98$, ratio $1.04$; our renders: $0.10$--$0.12$, $0.97$--$0.98$,
$1.00$--$1.04$) and the far field shifts by $-22$ and $+29\,\mathrm{px}$, against $-46$ and
$+36\,\mathrm{px}$ for ours. At weight $1.0$, the setting a fidelity criterion selected
beforehand (the smallest difference to the source outside the road), the profile has $3.3$ times
the authored height and drifts (RMSE $0.72\,\mathrm{m}$), and at $0.3$ it does not follow it
either (RMSE $0.38\,\mathrm{m}$, ratio $0.70$); the rule is met by neither
(Table~\ref{tab:depth_sweep}).

The weight-$0.6$ output
satisfies the camera-trajectory criterion but changes the surrounding appearance, with an outside-road
difference of $44.6/255$ against $4.5$--$15.5/255$ for the text-only outputs. The result shows
that a generator can texture a scaffold that already carries the authored geometry.

\paragraph{Models and settings.} All runs use one $81$-frame, $10\,\mathrm{fps}$,
$480\times320$ clip (the flat Scene~22 render), seed $2025$, and prompt enhancement
switched off. VACE is Wan2.1-VACE-14B with the source clip, a road-edit mask derived
from the input geometry,
and text; it runs $20$ UniPC steps at guidance scale $5.5$. Cosmos is Cosmos-Transfer2.5-2B
with its \emph{vis} control computed on the fly from the input clip: $35$ steps, guidance
$3$, control weight $0.60$ with medium blur. LTX is LTX-Video~0.9.8 13B in video-to-video mode with
its negative prompt, using the source-preserving preset (\texttt{preserve24}). None receives a target depth, mask of the target
shape, or reference frame. Pilot runs with smaller or inconsistent edits are not included
in the reported tables; all reported comparisons use the final coherent profiles and
settings described above.

\paragraph{Prompts.} The text given to Cosmos is below. VACE receives, in addition, the
sentence ``Apply the edit inside the supplied tracked ego-road mask. Preserve pixels and
identities outside the mask except for unavoidable geometric occlusion boundaries.''
before it, and LTX receives ``Perform a conservative video-to-video edit with strong
source identity preservation.'' All runs share the negative prompt at the end.
\begin{quote}\footnotesize
\textbf{Sag.} Forward-facing dashcam. Preserve the same street, buildings, vehicles,
route, and timing. Make one broad static sag from s=12 m to s=58 m: -8\% downhill, one
smooth low point at s=35 m, then +8\% uphill; not a pothole or trench. Make the broad
U-shaped vertical curvature clearly visible in road perspective, horizon placement, and
road-edge convergence. Grade 8\% means 8 m per 100 m, not 8 degrees. The edit is static
from frame 0 and fixed in the world: start frame 13 (1.30 s), center frame 41 (4.10 s),
end frame 78 (7.80 s). It never morphs or follows the car. Markings, curbs, wheels, and
vehicles follow the surface. Show natural camera heave and pitch with stable
perspective.

\textbf{Crest.} As for the sag, with ``Make one broad static crest from s=12 m to s=58 m:
+8\% uphill, one smooth high point at s=35 m, then -8\% downhill; not a short bump. Make
the broad inverted-U vertical curvature clearly visible \ldots''.

\textbf{Uphill / downhill.} As for the sag, with ``Make the ego road a static +8\%
uphill [-8\% downhill] from s=12 m to s=58 m in the travel direction. Make the sustained
slope clearly visible in road perspective, horizon placement, and road-edge
convergence.''

\textbf{Negative prompt.} animated road swelling, road morphing over time, moving hump,
hump following the car, pothole, black hole, trench, cliff, broken or floating lane
markings, floating or duplicated vehicles, melting wheels, unchanged flat-road camera
attitude through the edited feature, camera cut, camera teleportation, sideways route
drift, zoom, fisheye change, new vehicles, removed vehicles, building deformation,
weather change, lighting change, texture flicker, frame flicker, text, watermark.
\end{quote}

\paragraph{Frame anchors.} The prompts tie $s=12$, $35$, and $58\,\mathrm{m}$ to frames
$13$, $41$, and $78$ using the station map of the frozen model's flat run. The render's
own route station, integrated from the recorded camera path, places $s=36\,\mathrm{m}$ at
frame $45$ (frame $41$ is at $33.2\,\mathrm{m}$), about three frames later. The columns of
the figures follow the render's stations ($s=20$, $36$, and $52\,\mathrm{m}$), and every
generator tile is taken at the same frame index as our tile in its column.

\paragraph{Our arm.} The sag and crest use vertical curves of $16$, $32$, and
$16\,\mathrm{m}$ centerd on PVIs at $12$, $36$, and $60\,\mathrm{m}$ with grades
$\mp8\%$ to $\pm8\%$ (so $L/A=2.0\,\mathrm{m}$ per percent on each curve); the sustained
uphill and downhill enter and leave an $\pm8\%$ grade through $16\,\mathrm{m}$ curves at
$12$ and $58\,\mathrm{m}$ ($K=2.0$ as well). All of these are below the internal $K=4$
plausibility threshold used here and are stress geometry. Frames come from the
StreetUnveiler reconstruction deformed by the authored field. The sag, crest, and flat
frames are the earlier renders of this scene; the uphill and downhill frames are
rendered with the current code. A re-render of the sag with that code differs from the
earlier render by $1.4/255$ on average (at most $1.6/255$ in any frame), with a far-field
shift of $0.09\,\mathrm{px}$, so the two sets are interchangeable here.
The vehicle response is the frozen model run with the flat drive's recorded speed
profile $v_{\rm rec}(t)$ as the speed target, with total drive and brake torque
$T_{d}=\min\!\big(1200,\,900\,[v_{\rm rec}-v]_{+}\big)$ and
$T_{b}=\min\!\big(2000,\,900\,[v-v_{\rm rec}]_{+}\big)\,\mathrm{N\,m}$ (rear-wheel drive;
brake split $60/40$ front/rear; no comfort preview), $\mu=0.30$, and a $1\,\mathrm{ms}$ step; the flat rerun is
bit-identical to the earlier flat run.

\paragraph{Measurement.} The far-field shift is the median vertical optical flow
(Farneback, gray scale) over textured pixels above the horizon row of the flat render.
A median over the whole band is not used: the band contains large textureless sky
regions, whose flow is zero, so a median over all pixels is pulled towards zero on
frames where sky dominates. Correlation and slope are computed against $\arctan H'(s)$
at the render's route station, per road.

\begin{figure}[t]
\centering
\includegraphics[width=\linewidth]{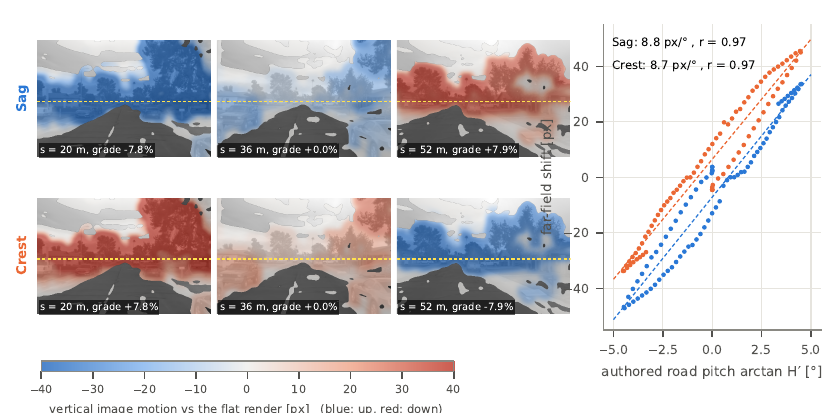}
\caption{Where our sag and crest are visible. Left: signed vertical image motion
against the flat render of the same frame, overlaid on the flat frame (grey: too smooth
to measure). The textured far field moves together. Right: far-field shift against
the authored road pitch, per frame; the fits are parallel lines whose offsets (about
$\mp7\,\mathrm{px}$) are consistent with the height parallax of a camera sitting up to
$1.28\,\mathrm{m}$ lower or higher.}
\label{fig:generators_visibility}
\end{figure}

\paragraph{Scene 22 sag and crest.} Figure~\ref{fig:generators} shows the text-only
comparison on the Scene~22 clip, the clip on which Table~\ref{tab:generators} is measured.

\begin{figure}[t]
\centering
\includegraphics[width=\linewidth]{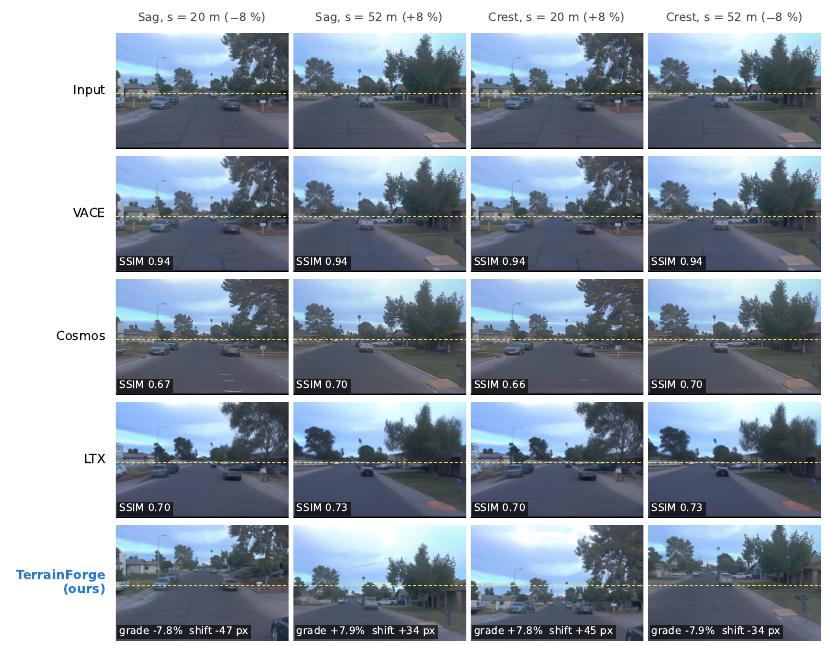}
\caption{Text-only comparison on the Scene~22 clip. Columns: the sag and the crest at
their steepest positions. Rows: the flat input; three generators that receive it with a
text description and no target geometry; and TerrainForge. The dashed yellow line is the
flat-road horizon, identical on every tile. Generator tiles show SSIM against the
input, which rewards not editing; our tiles show the authored grade and the far-field
shift against the flat render.}
\label{fig:generators}
\end{figure}

\begin{table}[t]
\centering
\caption{Outcome per case. Generators: what the output shows, and SSIM against the input
(mean over every fourth frame). \emph{copy}: the lane region stays within $1.25\times$ the
adjacent-frame difference of the input and the far field does not move;
\emph{regenerated}: the street content is replaced. Ours: far-field shift against the
flat render at $s=20$ and $52\,\mathrm{m}$, in pixels.}
\label{tab:generators}
\scriptsize
\begin{tabular}{l l l l l}
\hline
Case & VACE & Cosmos & LTX & Ours (px) \\
\hline
Sag $\pm8\%$ & copy, $0.94$ & regenerated, $0.69$ & regenerated, $0.73$ & $-47$ / $+34$ \\
Crest $\pm8\%$ & copy, $0.94$ & regenerated, $0.69$ & regenerated, $0.73$ & $+45$ / $-34$ \\
Uphill $+8\%$ & copy, $0.94$ & regenerated, $0.70$ & regenerated, $0.73$ & $+3$ / $+32$ \\
Downhill $-8\%$ & copy, $0.94$ & regenerated, $0.70$ & regenerated, $0.73$ & $-2$ / $-32$ \\
\hline
\end{tabular}
\end{table}

\paragraph{Conditioned arms: protocol.} The conditioned runs use the source clip, the road mask, and
the control videos of Waymo 150623, one run per cell at seed $2025$. Cosmos-Transfer2.5-2B runs $35$
steps with the depth or edge control at weight $1.0$ inside the road mask and the source latents
guided outside it for the \emph{road only} arms; for the \emph{full frame} arms the depth control
covers the whole frame, the first frame is the context, and nothing is guided. VACE runs $20$ steps
on a source video into which the lines or the depth are composited inside the road mask. The control
weights $0.3$, $0.6$, and $1.0$ are evaluated with a source-fidelity criterion: the smallest mean
absolute difference to the flat source outside the road mask, with ties within $0.5/255$ going to
the higher SSIM. The criterion does not use the road shape and selects $1.0$
(Table~\ref{tab:depth_sweep}). Fidelity to the source falls
monotonically as the weight is lowered, so there is no interior optimum. By eye, weight $0.3$ loses the
scene (vignette, a black disc, a different street), $0.6$ is a different street with invented markings
and a stop sign, and $1.0$ keeps the source layout but is over-saturated and painterly.

The camera-trajectory thresholds were calibrated on ten control sequences—three positive sag renders,
three flat renders, and four hard negatives—and then applied unchanged to generator outputs.
The controls separate positive camera trajectories from flat and non-sag edits; a near-copy of
the source also lands on the flat-render values (RMSE $0.52\,\mathrm{m}$, correlation $-0.24$).

\begin{table}[t]
\centering
\caption{Cosmos with depth over the whole frame, Waymo 150623 sag, one seed. Difference and SSIM are
against the flat source outside the ego-lane mask; the last three columns come from the recovered camera
height (the rule requires RMSE $\leq0.28$, correlation $\geq0.70$, ratio $0.6$--$1.4$); the last column is
the far-field shift at $s=20$ and $52\,\mathrm{m}$ in pixels. Ours: $-46$ and $+36$.}
\label{tab:depth_sweep}
\scriptsize
\begin{tabular}{l r r r r r l r}
\hline
Weight & Diff.\ /255 & SSIM & RMSE (m) & Corr. & Ratio & Rule & Shift (px) \\
\hline
$0.3$ & $64.8$ & $0.33$ & $0.38$ & $0.72$ & $0.70$ & not met & $-1$ / $-8$ \\
$0.6$ & $44.6$ & $0.41$ & $0.10$ & $0.98$ & $1.04$ & met & $-22$ / $+29$ \\
$1.0$ (selected) & $40.1$ & $0.42$ & $0.72$ & $0.75$ & $3.31$ & not met & $-6$ / $+36$ \\
\hline
\end{tabular}
\end{table}

\begin{figure}[t]
\centering
\includegraphics[width=\linewidth]{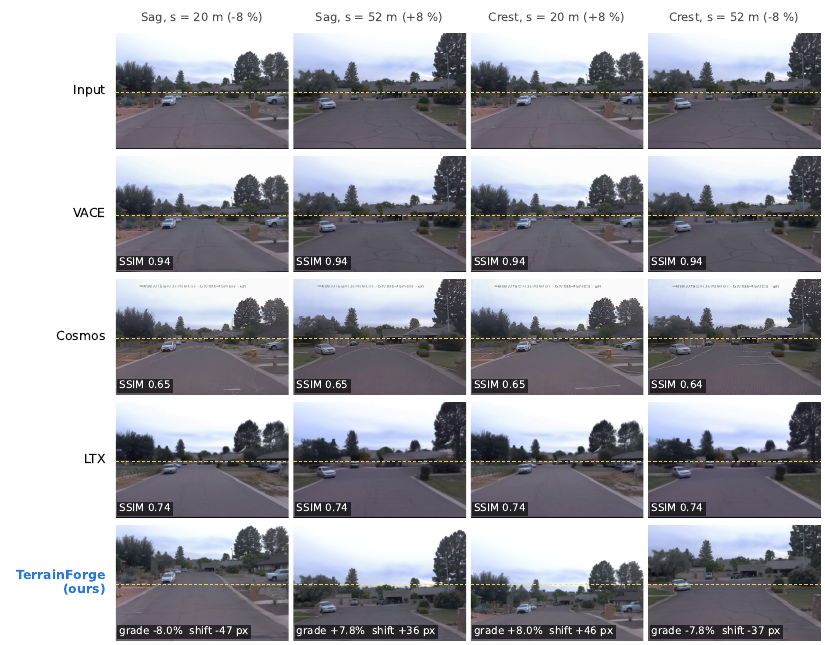}
\caption{Text-only comparison on Waymo 177619, in the layout of Figure~\ref{fig:generators}. The
columns are at $s=20$ and $52\,\mathrm{m}$.}
\label{fig:generators_177619}
\end{figure}

\begin{figure}[t]
\centering
\includegraphics[width=\linewidth]{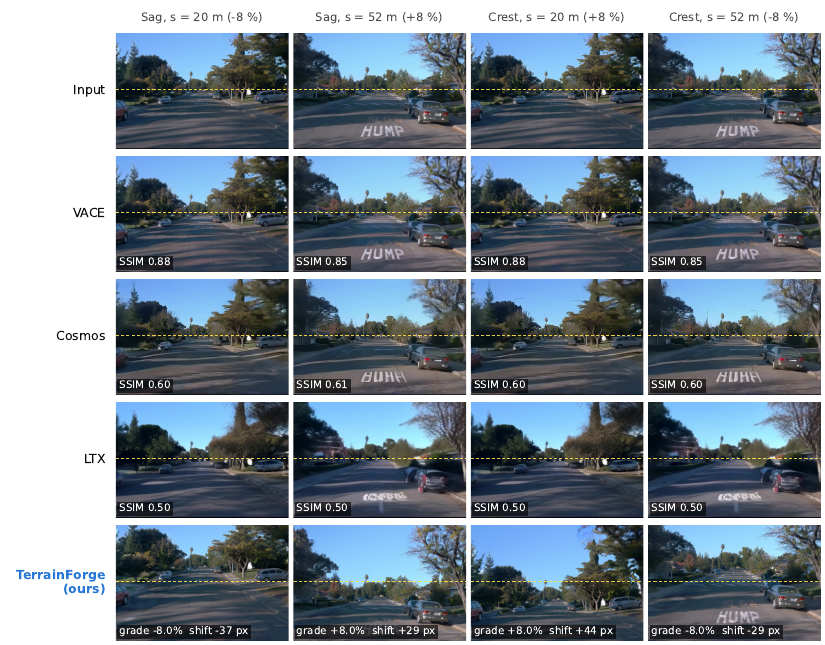}
\caption{Text-only comparison on Waymo 398895, in the layout of Figure~\ref{fig:generators}. The
columns are at $s=20$ and $52\,\mathrm{m}$.}
\label{fig:generators_398895}
\end{figure}

\paragraph{Further scenes.} Figure~\ref{fig:generators_177619} and Figure~\ref{fig:generators_398895}
show the text-only comparison on the other two further scenes, in the layout of Figure~\ref{fig:generators}.

\begin{table}[t]
\centering
\caption{Sag and crest ($\pm8\%$) on three further Waymo scenes. SSIM of each generator
against the input (mean over every fourth frame); largest far-field shift of any of the
three generators against the input over all frames; and our far-field shift against the
flat render at $s=20$ and $52\,\mathrm{m}$.}
\label{tab:generators_scenes}
\scriptsize
\begin{tabular}{l l r r r r r}
\hline
Scene & Case & VACE & Cosmos & LTX & Generators, max (px) & Ours (px) \\
\hline
150623 & Sag & $0.90$ & $0.72$ & $0.69$ & $1.4$ & $-46$ / $+36$ \\
 & Crest & $0.88$ & $0.69$ & $0.69$ & $1.5$ & $+44$ / $-36$ \\
177619 & Sag & $0.94$ & $0.66$ & $0.75$ & $1.2$ & $-47$ / $+36$ \\
 & Crest & $0.94$ & $0.65$ & $0.75$ & $1.1$ & $+46$ / $-37$ \\
398895 & Sag & $0.88$ & $0.62$ & $0.53$ & $1.8$ & $-37$ / $+29$ \\
 & Crest & $0.88$ & $0.61$ & $0.53$ & $1.9$ & $+44$ / $-29$ \\
\hline
\end{tabular}
\end{table}

\begin{figure}[t]
\centering
\includegraphics[width=\linewidth]{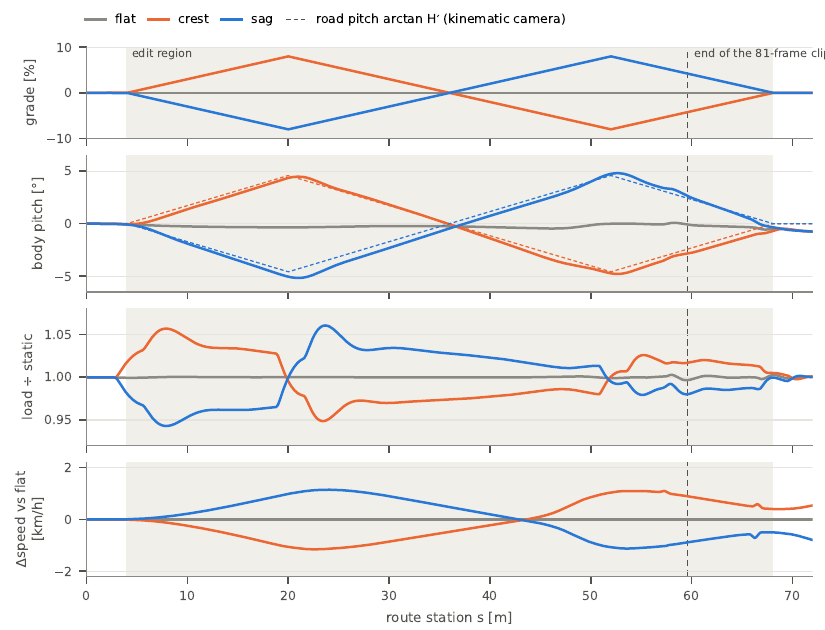}
\caption{Vehicle response of the frozen model to the authored sag and crest under the same
driver intent, plotted against route station. Solid: body pitch from the coupled model;
dashed: road pitch $\arctan H'$, which is what a kinematic camera sees. The gap between
them is the suspension response. Total normal load is the sum over the four wheels
divided by static weight. The generators have no such channel.}
\label{fig:generators_physics}
\end{figure}

\paragraph{Vehicle response of the sag and crest.} The generators output pixels only. In our arm the
same profiles give a body pitch of up to $5.2^\circ$ in magnitude against a road pitch
of $4.6^\circ$, a total normal load between $0.94$ and $1.06$ of the static weight, and
a speed difference from the flat road of about $\pm1\,\mathrm{km/h}$ under the same
driver intent (Figure~\ref{fig:generators_physics}). Load extremes occur a few meters
after the curvature steps at the ends of each vertical curve, where the suspension
transient rings; the speed effect is small because the driver tracks the recorded speed.

\begin{figure}[t]
\centering
\includegraphics[width=\linewidth]{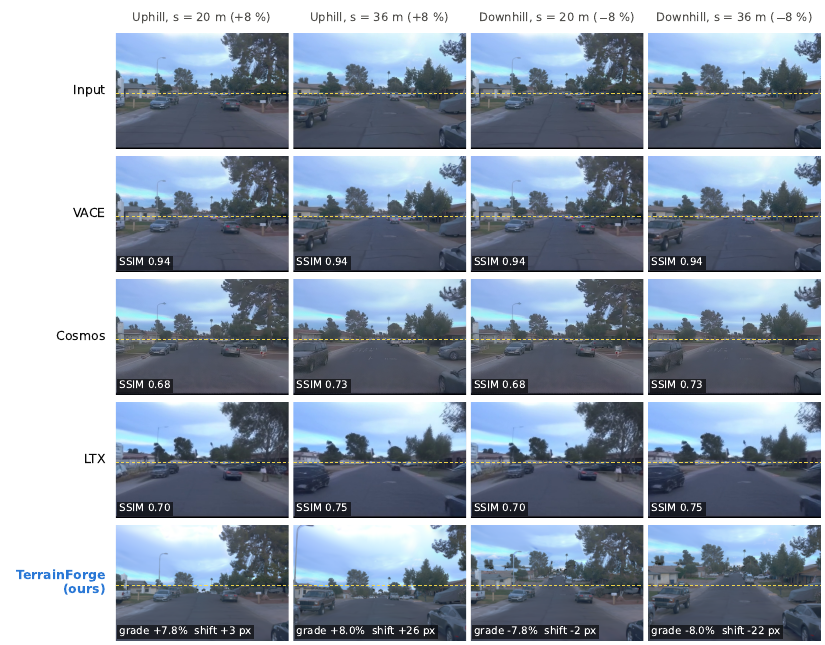}
\caption{The sustained $\pm8\%$ uphill and downhill, in the layout of
Figure~\ref{fig:generators}. The columns are at $s=20\,\mathrm{m}$, where the grade has
just been reached, and $36\,\mathrm{m}$.}
\label{fig:generators_grades}
\end{figure}

\begin{figure}[t]
\centering
\includegraphics[width=\linewidth]{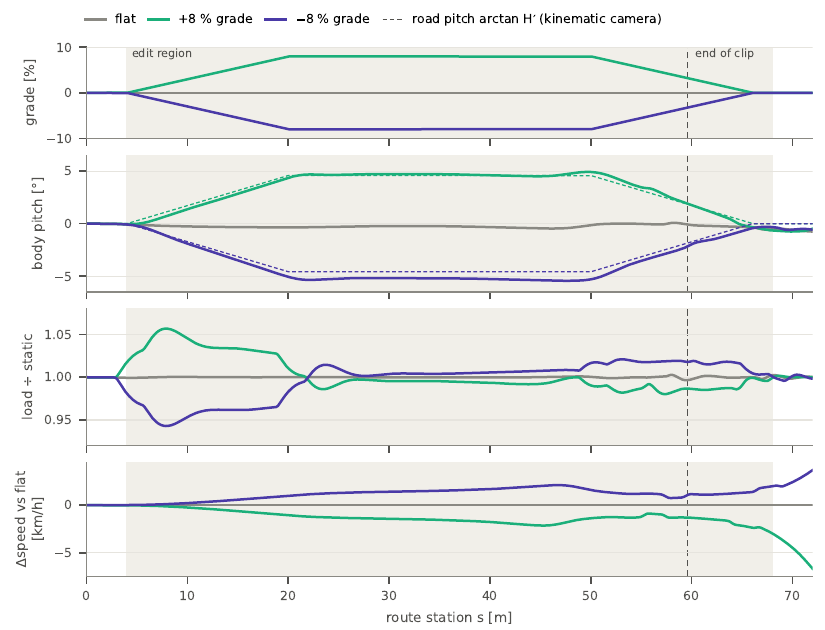}
\caption{Vehicle response to the sustained $\pm8\%$ grades, in the layout of
Figure~\ref{fig:generators_physics}. The speed axis is wider than there, and the plot is
cut at $s=72\,\mathrm{m}$.}
\label{fig:generators_physics_grades}
\end{figure}

\paragraph{Sustained grades.} Figure~\ref{fig:generators_grades} repeats the comparison
for the $\pm8\%$ uphill and downhill. The generators again copy or regenerate. Our
frames show why a sustained grade is a weak visual counterfactual: at $s=20\,\mathrm{m}$ the
grade is already $\pm7.8\%$ but the far field has moved by $+3$ and $-2\,\mathrm{px}$,
against $\pm45\,\mathrm{px}$ for the crest and sag at the same station, which is
consistent with the sheared streetscape rising with the road and cancelling the camera
pitch to first order. The
shift then grows to $+26$ and $-22\,\mathrm{px}$ at $36\,\mathrm{m}$ and $+32$ and
$-32\,\mathrm{px}$ at $52\,\mathrm{m}$, and it is no longer a linear function of road pitch
(slopes of $11.9$ and $10.4\,\mathrm{px}$ per degree, correlations $0.83$ and $0.80$),
because the camera has by then risen or fallen by up to $1.9\,\mathrm{m}$ at $s=36\,\mathrm{m}$
and the parallax is no longer small. The vehicle response is where the sustained grade
appears (Figure~\ref{fig:generators_physics_grades}): under the same driver intent the
speed falls up to $6.7\,\mathrm{km/h}$ below the flat road uphill and rises up to
$3.7\,\mathrm{km/h}$ above it downhill; body pitch
reaches $+4.9^\circ$ and $-5.4^\circ$ against road pitches of $\pm4.6^\circ$, and the
total normal load stays within $0.94$--$1.06$ of the static weight.

\section{Lateral response to a road-geometry edit: protocol and full results}
\label{app:lateral}

\paragraph{Protocol, fixed before any edited rollout.} The model is the coupled vehicle model of
Section~\ref{sec:carsim_validation}, and the pinned bank runner is imported unedited; the per-scene frozen
packages (initial $32$-state, wheel torque, steering, route geometry) are those of the friction bank, and
nothing is refitted per edit. The only change per arm is $\Delta H_{\eta}$ added to the prepared elevation
before the road surface is built, integrated at $0.5\,\mathrm{ms}$ with $5\,\mathrm{ms}$ output, identical
to the bank. Scenes are drawn at random (seed fixed in the protocol) among bank scenes whose baseline
covers at least $65\,\mathrm{m}$ of route station: $160$ \textsc{curve excitation} and $40$
\textsc{low excitation control}. The edit is a bounded symmetric crest or sag of total length
$60\,\mathrm{m}$ that returns to the original profile ($15\,\mathrm{m}$ entry, $30\,\mathrm{m}$ main curve,
$15\,\mathrm{m}$ exit), so $K=7.5$, $3.75$, and $1.875\,\mathrm{m}$ per percent for $g=2$, $4$, and $8\%$;
only $g=2\%$ meets the internal plausibility threshold used here. The apex is the station of peak $|a_y|$ of the baseline, clipped so
the whole edit lies ahead of the vehicle at $t=0$ and is fully traversed. An identity phase on $20$ scenes
reproduced the bank records exactly (all $11$ scalars, $|\text{diff}|=0$), which is what licenses using the
bank record as the baseline arm. Two amendments were made and recorded before any edited rollout was
scored: the apex rule was corrected because the global $|a_y|$ peak is the replay's initialisation
transient in $111$ of $200$ scenes, and the primary analysis was restricted to scenes whose baseline keeps
$v_x\geq1\,\mathrm{m/s}$ through the edit window, because the model cannot relaunch a vehicle brought to
rest. All $200$ scenes are run either way.

\paragraph{Full results.} Phase 1 ($\mu=0.8$) and phase 2 ($\mu=0.4$) each ran $1{,}200$ rollouts, all
complete; the primary set is $1{,}062$ runs over $177$ scenes ($142$ curve, $35$ control), with zero
stop-latch runs inside it. Table~\ref{tab:lateral_edit} gives the median paired deltas of the edit-window
peaks with $95\%$ scene-bootstrap intervals. Every quantity is monotone in $g$ at both friction levels.
The curve-versus-control separation appears in lateral acceleration and not in the load quantities, where
the control group is if anything larger and the intervals overlap; that is the mechanism rather than a
null, since unloading over a crest is set by $v^{2}H''/g$ and needs no lateral excitation. Sideslip and
yaw-rate-error medians are at the $10^{-4}$ level. Halving $\mu$ doubles the tire-utilisation change with
disjoint intervals and leaves the kinematic and load changes unchanged. Of the $1{,}200$ edited runs at
each friction level, $48$ runs raise at least one hard-failure flag. Across both phases this produces
$142$ flag instances, of which $141$ are \texttt{DOMAIN\_OR\_ROUTE\_SUPPORT\_EXIT}, meaning that the
edited run leaves the recorded route because its speed changes; exactly one run carries a tire-saturation
flag, at $\mu=0.4$. Flag counts
grow with $g$ and are the same at both friction levels, so they are a route-support boundary rather than a
physical one.

\begin{table}[t]
\centering
\caption{Lateral response to the road-geometry edit: median paired delta (edit minus baseline) of the
edit-window peak, with $95\%$ scene-bootstrap intervals for $g=8\%$. Curve: $142$ scenes; control: $35$.}
\label{tab:lateral_edit}
\scriptsize
\begin{tabular}{l l l r r l}
\hline
Quantity & Edit & Group & $2\%$ & $4\%$ & $8\%$ \\
\hline
Lateral acceleration [m/s$^2$] & crest & curve & $-0.0041$ & $-0.0081$ & $-0.0156$ [$-0.0203$, $-0.0130$] \\
 & crest & control & $-0.0024$ & $-0.0043$ & $-0.0089$ [$-0.0122$, $-0.0047$] \\
 & sag & curve & $+0.0041$ & $+0.0085$ & $+0.0180$ [$0.0149$, $0.0228$] \\
 & sag & control & $+0.0024$ & $+0.0049$ & $+0.0100$ [$0.0074$, $0.0127$] \\
Minimum load ratio & crest & curve & $-0.0217$ & $-0.0543$ & $-0.1309$ [$-0.1477$, $-0.1041$] \\
 & crest & control & $-0.0250$ & $-0.0641$ & $-0.1466$ [$-0.1986$, $-0.0920$] \\
Tire utilisation, $\mu=0.8$ & sag & curve & $+0.0012$ & $+0.0034$ & $+0.0094$ [$0.0066$, $0.0125$] \\
Tire utilisation, $\mu=0.4$ & sag & curve & $+0.0025$ & $+0.0068$ & $+0.0192$ [$0.0133$, $0.0253$] \\
\hline
\end{tabular}
\end{table}

\section{Additional screening results}
\label{app:screening}

\paragraph{Lateral bank.} On the lateral bank, the better of two trajectory-based learners (a dual-resolution sequence model and summary-statistics trees) reaches $R^2$ of $0.76$--$0.98$ for ten of the eleven peak quantities and $0.58$ for path deviation, while static scene descriptors reach at most $0.58$, so the information is in the trajectory (Table~\ref{tab:lateral_screening}, Appendix~\ref{app:screening}). The sequence model is significantly better than the summary learner on five quantities, worse on two, and indistinguishable on four.

\paragraph{Extended description of the screening study (from the main text).}
\paragraph{Banks.} Both banks use a 70/15/15 scene-disjoint split fixed before any
outcome was computed, and each was drawn twice, as two disjoint sets of $1{,}000$
episodes selected by the same rules. The \emph{obstacle} bank contains episodes in
which the ego vehicle brakes toward a recorded vehicle, pedestrian, or cyclist. Only the ego is re-simulated; the target is replayed from its recording and nothing is rendered. The
second draw adds vehicle targets only, because the first used every qualifying
pedestrian and cyclist episode. Each episode receives compact SAG edits (central
length $12$ or $20\,$m, grade $2$ or $6\%$, early or late in the braking corridor) and
a $6\,$cm, $3.66\,$m raised-cosine hump, at $\mu\in\{0.4,0.8\}$. Of the $2{,}000$
episodes, $1{,}221$ pass a strict target-semantics gate and $983$ contribute complete
action rows, giving $15{,}758$ edits ($11{,}834$ SAG, $3{,}924$ hump) over
$698/143/142$ train/validation/test scenes. Outcomes are the changes in the ego--target
minimum signed clearance $d_{\min}$ and in the terminal gap $G$ relative to the unedited road (Appendix~\ref{app:method}).
Although the episodes are selected for braking, the replayed ego comes to a stop within the horizon in only $8.5\%$ of the edited rollouts; the others end at a median of $6.7\,\mathrm{m/s}$, and both arms of a comparison stop in $7.0\%$ of the rows. The obstacle episodes are
straight-line braking and were simulated with the earlier front lateral tire-force
scale, which differs from the final one only under steering
(Section~\ref{sec:carsim_validation}). The \emph{lateral} bank contains episodes on
the recorded road ($800$ curve-excitation and $200$ low-excitation controls per draw)
at $\mu\in\{0.4,0.6,0.8\}$, simulated with the final model, giving $3{,}123$ valid
rollouts over $714/152/178$ scenes. Its targets are eleven peak lateral and load
quantities per rollout. The lateral bank varies friction on the recorded road; it does
not contain road edits.

\paragraph{Physics prior.} For the obstacle bank we add a closed-form estimate that
reads no simulator output. If the replayed actuation applies, to first order, the
same retarding force as a function of station on both roads, the edit changes the
speed only through the elevation it adds:
\begin{equation}
v_{\eta}^{2}(s)=v_{\rm rec}^{2}(s)-2g\,\big[\Delta H_{\eta}(s)-\Delta H_{\eta}(s_0)\big],
\label{eq:energy_prior}
\end{equation}
where $v_{\rm rec}$ is the recorded speed. Integrating $\mathrm{d}t=\mathrm{d}s/v_{\eta}$
to the recorded horizon, or to the first zero of $v_{\eta}^2$, gives the station
shift and hence a prior on $\Delta G$. We also supply the speed at
the edit, the hump's vertical acceleration $(h/2)(2\pi/L)^2v^2$, and the SAG
curvature load $v^2\kappa/g$. The prior covers the gravitational part of the edit
only; wheel unloading over a hump is not in it.

\paragraph{Protocol.} Hyper-parameters are chosen on the validation split and the
test split is scored once; tree and linear models are then refit on train and
validation, and the lateral sequence model keeps its validation-selected
checkpoint. Every baseline is fit on the same rows with the same split. For the
obstacle bank the evaluation protocol, including every model, baseline, and metric
below, was written down before any model was fit on the combined banks. Intervals are
$95\%$ bootstrap intervals over whole test scenes ($2{,}000$ resamples).

\begin{table}[t]
\centering
\caption{Predicting the effect of a candidate edit on held-out scenes, all targets and baselines. MAE in meters over all
test edits; ratio is MAE over that of predicting no change; $\rho$ is the Spearman rank
correlation between predicted and simulated change, pooled over test edits of all scenes.
Brackets: $95\%$ scene-bootstrap intervals. Test edits: $1{,}684$ sag ($109$ scenes), $568$ hump
($142$ scenes).}
\label{tab:screening_full}
\small
\setlength{\tabcolsep}{4pt}
\begin{tabular}{llccccc}
\toprule
Target & Edit & No change & Closed-form prior & Boosted trees & MAE ratio & $\rho$ \\
\midrule
$\Delta G$ & sag  & $0.451$ & $0.469$ & $0.270$ & $0.60$ [$0.41$, $0.76$] & $0.89$ [$0.78$, $0.96$] \\
$\Delta G$ & hump & $0.730$ & $1.043$ & $0.569$ & $0.78$ [$0.44$, $0.89$] & $0.81$ [$0.72$, $0.89$] \\
$\Delta d_{\min}$ & sag  & $0.028$ & $0.097$ & $0.027$ & $0.97$ [$0.85$, $1.06$] & $0.21$ [$0.05$, $0.37$] \\
$\Delta d_{\min}$ & hump & $0.363$ & $0.634$ & $0.360$ & $0.99$ [$0.91$, $1.02$] & $0.30$ [$0.14$, $0.44$] \\
\bottomrule
\end{tabular}
\end{table}

\paragraph{Obstacle results.} Table~\ref{tab:obstacle_screening} gives the effect-detection result.
An edit counts as effective if it creates a collision or moves $d_{\min}$ by
$0.1\,$m or $G$ by $0.5\,$m. Of the $2{,}673$ effective edits, $20$ involve a collision, $796$ move $d_{\min}$, and $1{,}873$ qualify through $G$ alone. Road geometry alone is a weak
predictor of this (AP $0.35$); the scene context (ego speed, target distance,
braking strength) is what makes it predictable (AP $0.85$, against $0.20$ at random).
Reviewing the top-ranked fifth of the edits recovers $80\%$ of the effective ones. The
trees are better than a logistic model on the same features, but only modestly (paired
$\Delta$AP interval [$0.04$, $0.15$]), against [$0.42$, $0.58$] over geometry-only
logistic, so most of the gain comes from the features rather than the learner. Detection
improves steadily with data: trained on $25$, $50$, $75$, and $100\%$ of the training
scenes, AP is $0.78$, $0.81$, $0.83$, and $0.85$, with no plateau yet.

\begin{table}[t]
\centering
\caption{Obstacle-intervention effect detection on the scene-disjoint test split. AP and
Recall@20\% are reported for the same labels used in the main text.}
\label{tab:obstacle_screening}
\small
\begin{tabular}{lcc}
\hline
Model & AP & Recall@20\% \\
\hline
Random ranking & 0.20 & 0.23 \\
Geometry-only logistic & 0.35 & 0.44 \\
Full-context logistic & 0.75 & 0.75 \\
Boosted trees & 0.85 & 0.80 \\
\hline
\end{tabular}
\end{table}

The change in $G$ is predictable in rank and, on average, in size: the
error is $22$--$40\%$ below predicting no change, with rank correlation $0.81$--$0.89$.
We report MAE ratios and rank correlation rather than $R^2$, because the deltas are
heavy-tailed: the top $5\%$ of test rows carry $48\%$ (SAG) and $75\%$ (hump) of the
total absolute change. For the same reason the loss matters. The protocol also included
a second regressor with a fixed squared-error loss, and it does not beat predicting no
change (MAE ratio $1.27$ for SAG and $1.49$ for the hump), because squared error chases
the tail; the boosted trees select their loss on the validation split and chose
absolute error in every fit. The physics prior of Eq.~\ref{eq:energy_prior}, with no
learning and no simulation, ranks SAG edits at $0.68$ but does not get their size right
(MAE ratio $1.04$), and it carries no information about the hump, whose effect comes
from wheel unloading rather than elevation. As a feature it matters for SAG where an edit
actually changes the outcome: on those rows it raises rank correlation from $0.47$ to
$0.69$ (paired interval [$0.07$, $0.37$]); for the hump it makes no significant
difference ($0.15$ to $0.17$, [$-0.21$, $0.22$]). The clearance change is not
predicted better than no change: most edits leave the closest approach unchanged, and no model beats
predicting no change. More data does not change the regression much: in a scaling run
made after the protocol results were seen, the MAE ratio for $\Delta G$ moves from
$0.71$ to $0.66$ (SAG) and $0.79$ to $0.78$ (hump) between $25$ and $100\%$ of the
training scenes.

\paragraph{Lateral results.} Table~\ref{tab:lateral_screening} (Appendix~\ref{app:screening}) compares a
dual-resolution sequence model ($20\,$ms and $5\,$ms branches) with two simpler
learners. The response is predictable from pre-simulation inputs: under the better
of the two trajectory-based learners, $R^2$ is $0.76$--$0.98$ for ten of the eleven
quantities and $0.58$ for path deviation. The information is in the trajectory: the
static scene descriptors reach at most $0.58$, and a one-feature physics proxy at most
$0.38$. The steady-state formula is weak here because the excitation is small (median
peak lateral acceleration $0.29\,\mathrm{m/s^2}$). The sequence model is significantly
better than the summary learner on five quantities (yaw-rate error, lateral
acceleration, roll rate, and both wheel-load quantities), significantly worse on two
(sideslip and tire utilisation), and indistinguishable on four, so the architecture
helps for some quantities and not others. It also gains most from data: from $25$ to
$100\%$ of the $714$ training scenes its $R^2$ rises by $0.07$--$0.21$ on every
quantity. Tire utilisation is uncorrelated with lateral demand in this bank
($R^2=0.00$) and tracks peak wheel torque instead ($R^2=0.60$ for that single
feature), which the summary features expose directly.

Table~\ref{tab:lateral_screening} gives the per-quantity test accuracy of the lateral screening models of
Section~\ref{sec:screening}.

\begin{table}[t]
\centering
\caption{Lateral screening, $R^2$ on $533$ test rollouts from $178$ scenes, with
$95\%$ scene-bootstrap intervals. Static: scene-level descriptors only.
Summary: the sequence model's own pre-simulation inputs reduced to per-rollout
maximum, $95$th percentile, and mean.}
\label{tab:lateral_screening}
\scriptsize
\begin{tabular}{l r r r}
\hline
Peak quantity & Static trees & Summary trees & Sequence model \\
\hline
Sideslip angle & $0.06$ & $0.97$ [$0.92$, $0.99$] & $0.92$ [$0.85$, $0.95$] \\
Heading error & $-0.06$ & $0.62$ [$0.19$, $0.83$] & $0.76$ [$0.64$, $0.86$] \\
Yaw-rate error & $0.20$ & $0.75$ [$0.40$, $0.92$] & $0.94$ [$0.90$, $0.97$] \\
Lateral acceleration & $0.25$ & $0.72$ [$0.56$, $0.81$] & $0.89$ [$0.84$, $0.92$] \\
Roll angle & $0.58$ & $0.92$ [$0.88$, $0.95$] & $0.95$ [$0.92$, $0.96$] \\
Roll rate & $0.29$ & $0.65$ [$0.40$, $0.81$] & $0.87$ [$0.82$, $0.91$] \\
Load-transfer ratio & $0.51$ & $0.89$ [$0.84$, $0.93$] & $0.92$ [$0.89$, $0.94$] \\
Minimum wheel load & $0.18$ & $0.69$ [$0.58$, $0.78$] & $0.90$ [$0.86$, $0.93$] \\
Minimum load ratio & $0.15$ & $0.72$ [$0.60$, $0.81$] & $0.90$ [$0.85$, $0.93$] \\
Tire utilisation & $0.35$ & $0.98$ [$0.97$, $0.99$] & $0.91$ [$0.87$, $0.94$] \\
Path deviation & $0.26$ & $0.58$ [$0.45$, $0.71$] & $0.52$ [$0.30$, $0.65$] \\
\hline
\end{tabular}
\end{table}

\end{document}